\documentclass[11pt]{article}

\usepackage[final]{acl}

\usepackage{times}
\usepackage{latexsym}

\usepackage[T1]{fontenc}
\usepackage[utf8]{inputenc}

\usepackage{microtype}

\usepackage{inconsolata}

\usepackage{enumerate}
\usepackage{amsmath,amssymb,amsfonts}
\usepackage{algorithmic}
\usepackage{graphicx}
\usepackage{textcomp}
\usepackage{xcolor}
\usepackage{enumitem}
\definecolor{green}{RGB}{0, 102, 0}

\usepackage{subcaption}
\usepackage{booktabs}
\usepackage{tabularx}
\usepackage{makecell}
\usepackage{float}
\usepackage{tcolorbox}
\usepackage{stfloats}

\usepackage{cleveref}
\title{Beyond Multiple Choice: Evaluating Steering Vectors for Summarization\thanks{Published in Findings of EACL 2026.}}
\author{
\textbf{Joschka Braun\textsuperscript{1}},
\textbf{Carsten Eickhoff \textsuperscript{1}},
\textbf{Seyed Ali Bahrainian\textsuperscript{1}}
\\
\textsuperscript{1} University of Tübingen
\\
  \small{
    \textbf{Correspondence:} \href{mailto:joschkacbraun@gmail.com}{joschkacbraun@gmail.com}
  }
}

\begin{document}
\maketitle
\begin{abstract}
Steering vectors are a lightweight method for controlling text properties by adding a learned bias to language model activations at inference time. While predominantly studied for multiple-choice and toy tasks, their effectiveness in free-form generation remains largely unexplored. Moving ``Beyond Multiple Choice,'' we evaluate steering vectors for controlling topical focus, sentiment, toxicity, and readability in abstractive summaries across the SAMSum, NEWTS, and arXiv datasets. We find that steering effectively controls targeted properties, but high steering strengths consistently induce degenerate repetition and factual hallucinations. Prompting alone preserves summary quality but offers weaker control. Combining both methods yields the strongest control and the most favorable efficacy-quality trade-off at moderate steering strengths. Our work demonstrates that steering vectors face a critical control-quality trade-off in free-form generation, and that hybrid approaches offer the best balance in practice.
\end{abstract}

\section{Introduction}
Large pre-trained language models, trained on extensive web-based datasets, have become the standard approach for a wide range of tasks in natural language processing~\cite{BERT_model, GPT3_LMs_are_Few-Shot_Learners}. Consequently, the ability to adapt foundation models to specific tasks and align their outputs with user preferences has become crucial. Recent methods for controlling language models can often be classified into three main strategies: prompt engineering~\cite{AutoPrompt_Eliciting_Knowledge_Shin_2020, Lester_Parameter_Efficient_Prompt_Tuning, Chain_of_thought_prompting}, trainable decoding mechanisms~\cite{Residual_Energy_Based_Models_trainable} and fine-tuning according to specific objectives~\cite{RLHF, Direct_Preference_Optimization}. However, prompting often provides insufficient control, while alternatives require significant engineering, computation, or data.

\noindent A promising fourth strategy is \textit{activation engineering}, which directly modifies model activations during text generation~\cite{Representation_Engineering_Dan_Hendrycks}. For example, \textit{difference of means steering vectors}~\cite{Steering_Llama2_via_Contrastive_Activation_Addition} compute a direction in the activation space by taking the mean of activations from examples of a target behavior and subtracting the mean of activations from counter-examples. This resulting steering vector can then be added to the model's activations during inference to align outputs with user preferences. Although previous research demonstrates the effectiveness of steering methods in multiple-choice settings and simplified toy tasks, their practical effectiveness for free-form generation tasks remains understudied. We address this gap by evaluating steering vectors for adaptive summarization on three diverse abstractive summarization datasets covering conversations, news, and scientific articles. Through six research questions, we examine the efficacy-quality trade-off and provide actionable guidelines for effective steering.\vspace{2pt}\\
\noindent\textbf{Our contributions are:}\vspace{2pt}
\begin{enumerate}[leftmargin=1.5em, itemsep=1.5pt, topsep=0pt, parsep=0pt]
    \item We evaluate steering vectors to control topical focus, sentiment, toxicity, and readability in adaptive free-form summaries on diverse datasets. Except for toxicity, all text properties can be effectively induced.
    \item We evaluate generated summaries for unwanted side effects,
    finding that high steering strengths consistently
    induce degenerate repetition and factual hallucinations.
    \item We compare steering, prompting, and their combination: prompting alone preserves quality but offers weaker control, while the hybrid approach yields the strongest control and most favorable efficacy-quality trade-off at moderate steering strengths.
    \item We release our code (MIT) and datasets (CC BY-NC 4.0) to support reproducibility.\footnote{\href{https://github.com/JoschkaCBraun/adaptive-steering}{https://github.com/JoschkaCBraun/adaptive-steering}}
\end{enumerate}
\newpage
\section{Related Work}
\paragraph{LLM-based controllable summarization.} Generating adaptive summaries tailored to user preferences typically involves fine-tuning existing foundation models, modifying model architectures, or employing specialized training procedures~\cite{Controllable_Text_Summarization_Survey, text_simplification_via_adaptive_teaching_ali, braun2025logitreweightingtopicfocusedsummarization,Survey_on_process-oriented_ats_with_LLMs}.
For instance,~\cite{Blinova_Ali_SIMSUM} proposes a two-stage model for document-level text simplification that first summarizes and then further simplifies content using transformers, enhanced by keyword prompts and an embedding similarity loss.
\paragraph{Steering vectors for LLM control.} Controlling text generation by adding a steering vector is easier to implement and only requires sufficient training data to be effective. Steering vectors leverage the interpretability-based insight that many human-interpretable text properties like truthfulness~\cite{Geometry_of_Truth_Linear_Max_Tegmark, Inference-Time_Intervention_Wattenberg}, refusal~\cite{Refusal_Linear_direction_neel_Nanda} and sentiment~\cite{Activation_Addition_Turner, Sentiment_is_linear_Neel_Nanda} are likely represented linearly. Various methods based on this insight have been proposed to control LLM outputs~\cite{Extracting_Steering_Vectors_Subramani, Activation_Addition_Turner, Steering_Llama2_via_Contrastive_Activation_Addition, Inference-Time_Intervention_Wattenberg, In-Context_learning_tasks_Vectors, Function_Vectors_in_LLMs_David_Bau, Style_Vectors, Representation_Engineering_Dan_Hendrycks}.
\paragraph{Limitations of steering vectors.} Despite their appeal as lightweight control methods, activation steering methods face significant challenges~\cite{braun2024soberlookatsteeringvectorsinLLMs}. Recent studies highlight issues with reliability and generalization, noting high variance across inputs and instances where steering produces the opposite of the intended effect~\cite{Analyzing_the_Generalization_and_Reliability_of_Steering_Vectors_Daniel_Tan, Brumley_unreliability_of_steering, braun2025understanding_unreliability}.
Furthermore, steering vectors are often evaluated in constrained settings, like multiple-choice questions or simple toy tasks, rather than more challenging free-form generation tasks~\cite{Reliable_Evaluation_Itamar, braun2025beyond}.
\\
\\
\noindent Our work addresses this research gap by evaluating ``difference of means'' steering vectors for adaptive free-form summarization, testing the method across multiple language model families and parameter sizes on three diverse, established summarization datasets to assess its control efficacy and side effects for this common application.

\section{Research Questions}
\textbf{Q1. How effective are steering vectors for controlling free-form summarization?}
We evaluate the extent to which steering vectors can control summary properties like topical focus, sentiment, toxicity, and readability in abstractive summaries. We also examine how a model's safety training affects the ability to steer it towards suppressed behaviors like toxicity.
\newline\textbf{Q2. What are the side effects of steering on summary quality?}
We measure the impact of steering strength $\lambda$ on intrinsic and extrinsic summary quality to characterize the trade-off between control and quality degradation. We also probe for entanglement between the steering directions.
\newline\textbf{Q3. How does steering compare to prompting for text control?}
We compare the control efficacy and quality side effects of steering vectors against prompting for the same target behavior. This establishes a performance baseline for activation engineering versus a conventional control method.
\newline\textbf{Q4. How effective is a hybrid steering and prompting approach?}
We test if combining steering vectors with instructive prompts yields stronger control than either method alone. We also evaluate whether this hybrid approach offers a more favorable trade-off between control efficacy and quality degradation.
\newline\textbf{Q5. Can conditional steering mitigate steering side effects while maintaining control efficacy?} We investigate whether adjusting the steering strength $\lambda$ based on the model's current activations can mitigate quality degradation without sacrificing control over the target property.
\newline\textbf{Q6. How do model scale and architecture impact steering?}
To understand the generalizability of our findings, we examine how steering efficacy and its associated side effects scale with the number of parameters and layers in a language model. We also explore whether performance differs significantly across distinct transformer architectures, using models from the Llama, Qwen, and Gemma families.

\section{Methods and Experimental Setup}
\subsection{Datasets}
We evaluate the trade-off between control efficacy and text quality when applying steering vectors to three datasets selected for their diversity in domain, length, and linguistic complexity. SAMSum~\cite{samsum_dataset} consists of approximately 16,000 short, informal messenger-like dialogues. Moving up in length and complexity, NEWTS~\cite{NEWTS_dataset} contains 3,000 news articles derived from CNN/DailyMail~\cite{CNN_Daily_Mail_dataset}, specifically designed with dual reference summaries for topic-focused summarization. Finally, the arXiv dataset~\cite{arXiv_dataset} represents the most challenging domain, featuring long, highly technical scientific papers.

\subsection{Steering Method}
We use difference of means steering vectors by~\citet{Steering_Llama2_via_Contrastive_Activation_Addition} as the steering method. To compute the layer- and behavior-specific steering vector \(\mathbf{s}^l \in \mathbb{R}^{d}\) from training dataset \(\mathcal{D}_{\text{train}} = \{(x_i^+, x_i^-)\}^{N_{\text{train}}}_{i = 1}\), we record residual stream activations at layer $l$. 
Activations are recorded at the last position of the training sample. The resulting activations are denoted \(\mathbf{a}^l(x_i^+)\) and \(\mathbf{a}^l(x_i^-)\) respectively. 
The steering vector \(\mathbf{s}^l \in \mathbb{R}^{d}\) is the mean difference between positive and negative activations:
$
\mathbf{s}^l  = 1/|\mathcal{D}_{\text{train}}| \sum_{\mathcal{D}_{\text{train}}} \bigl[ \mathbf{a}^l(x_i^+) - \mathbf{a}^l(x_i^-) \bigr].
$
To steer during inference, we add $\lambda \mathbf{s}^l$ to the residual stream at layer $l$. 
Here $\lambda \in \mathbb{R}$ is the steering strength. Most of our experiments are done with a range of steering strengths, choosing $\lambda \in \{-5, -2, -1.5, -1, -0.5, 0, 0.5, 1, 1.5, 2, 5\}$.

\subsection{Evaluation Framework}\label{sec:evaluation_of_summaries}
We evaluate generated summaries by measuring both their overall quality and the efficacy of steering control. Summary quality is assessed from two perspectives. Intrinsic quality is measured by Perplexity (PPL)~\cite{Perplexity_Bengio} and textual diversity using word and character bigram repetition (Distinct-2)~\cite{N-Gram_repetition_paper}. Extrinsic quality is evaluated against reference summaries using ROUGE-1, -2, and -L~\cite{ROUGE_score} and semantic similarity with BERTScore~\cite{BERT_score, Deberta_model}. Control Efficacy is quantified for four target properties. We measure topical focus using LDA-based methods, sentiment with both lexicon-based (VADER)~\cite{VADER} and transformer-based classifiers~\cite{nlptown_bert_sentiment}, and toxicity using two separate transformer models~\cite{RoBERTa}. Finally, readability is scored using regression models based on DistilBERT~\cite{DistilBERT} and DeBERTa-V3~\cite{DeBERTav3}. We provide detailed metric descriptions and validate their correlation with LLM-based judgments in Appendix~\ref{app:sec:evaluation_of_summaries}.

\subsection{Prompt Engineering}
\label{sec:prompt_engineering}
We use a consistent prompt structure for all models and steering vectors in our primary experiments. The basic prompt $x$ is designed to elicit a general, neutral three-sentence summary and is formatted as follows:
{\fontsize{10.5}{11}\selectfont\begin{verbatim}
Write a {N} sentence summary of the article.
Article:
{article}
Summary:
\end{verbatim}
}
In this template, \texttt{{article}} is the placeholder for the input text, and the target sentence count \texttt{{N}} is set to one for the SAMSum dataset, three for NEWTS, and ten for the arXiv dataset. We defer the detailed description of prompt variations engineered to encourage or discourage selected text properties to the Appendix~\ref{app:prompt_variations}.

\subsection{Language Models}
To study the generalizability of our results and the influence of model scale, we experiment with thirteen models from three distinct families: Llama 3~\cite{Llama3_model} (1B, 3B, 8B, 70B), Qwen 3~\cite{Qwen3_model} (0.6B, 4B, 8B, 14B, 32B), and Gemma 3~\cite{Gemma_model} (1B, 4B, 12B, 27B). This selection allows us to assess how results generalize across different architectures and systematically analyze the impact of model size on steering efficacy.

\subsection{Summary Generation}
For summary generation, we set the maximum output length based on the characteristic length of the source documents and reference summaries for each dataset: 50 tokens for SAMSum, 150 for NEWTS, and 300 for arXiv. Unless otherwise specified, we evaluated each setting on a random sample of 250 articles from the respective dataset. As this data was not used for training the steering vectors, no data leakage occurs.

\subsection{Steering Setup} Steering was applied to middle-to-late layers of each model, a strategy that aligns with established heuristics and previous literature. The exact layer configurations for each model are available in our \href{https://github.com/JoschkaCBraun/adaptive-steering}{code repository}.
For training the steering vectors, we used custom datasets, each containing 100 contrastive samples of the desired behavior and its opposite. These samples were text pairs designed to primarily differ only in the target attribute (sentiment, toxicity, readability, or topic) and can be found in the same repository.

\newpage
\section{Results}
\subsection{Steering Vectors Successfully Control Target Behaviors}
\begin{figure}[!htp]
  \includegraphics[width=\columnwidth]{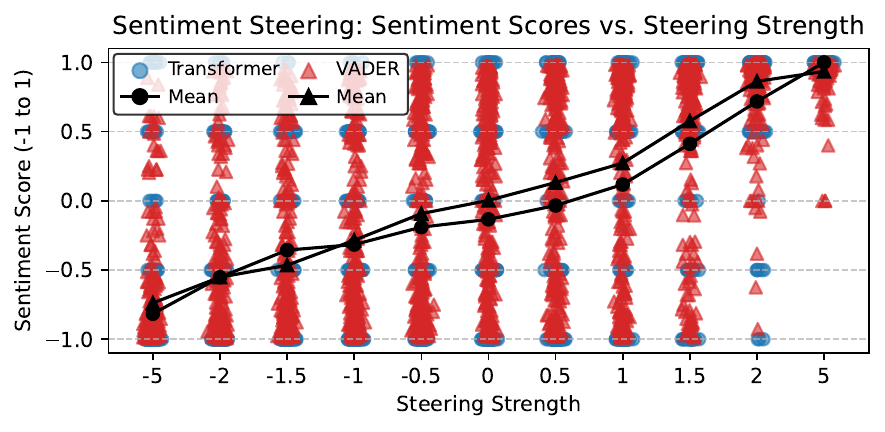}
  \caption{Steering vectors successfully control the sentiment of generated summaries in Llama 8b. Without steering the average sentiment is neutral. Negative and positive steering strength effectively shift the average sentiment towards the target polarity. Both metrics result in similar sentiment scores and measure a monotonic increase in sentiment relative to the applied steering strength.}
  \label{fig:sentiment_steering_impact_on_sentiment}
\end{figure}

We find that topical focus, sentiment (see \Cref{fig:sentiment_steering_impact_on_sentiment}), and readability can be effectively controlled across all datasets and model sizes.

\begin{figure}[!htp]
  \includegraphics[width=\columnwidth]{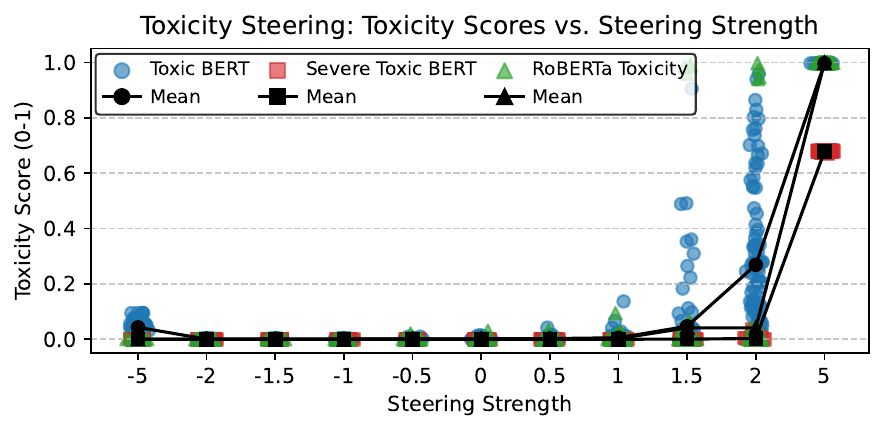}
  \caption{Steering for toxicity only impacts toxicity for steering strengths of 2 and larger. The safety-tuned Llama 8b model is able to avoid generating toxic text until very high steering strengths likely shift the activations out of distribution, bypassing post-training and massively degrading text quality.}
  \label{fig:toxicity_steering_impact_on_toxicity}
\end{figure}

However, toxicity is not elicited in the generated text except at high steering strengths (typically $\lambda > 2$, see \Cref{fig:toxicity_steering_impact_on_toxicity}), an outcome likely attributable to the model's safety training. This makes toxicity a compelling case study for examining the challenge of steering a behavior that the model has been explicitly trained to avoid.

\noindent More detailed steering results are in Appendix~\ref{app:impact_of_model_scale}. Complementing quantitative metrics from the results section, Appendix~\ref{app:individual_samples} provides qualitative summaries illustrating the impact of applied methods on text properties.

\begin{figure}[!htp]
   \includegraphics[width=\columnwidth]{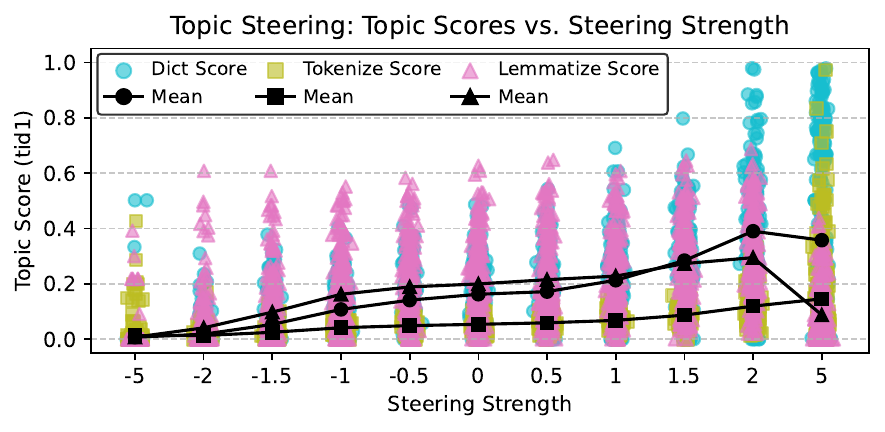}
   \vspace{-0.7cm}
   \caption{The topic scores for all three metrics increase monotonically for steering strengths up to 2. The effect size of steering strengths between -1 and 1 is relatively small, and there is a noticeable improvement for steering strengths larger than magnitude 1. Applying the vector with a negative factor makes the topic less dominant. For a steering strength of 5 the text degrades and the topic scores with it.}
  \label{fig:topic_steering_impact_on_topical_focus}
  \vspace{-0.6cm}
\end{figure}
\subsection{Steering Side Effects on Unrelated Properties}
\begin{figure}[!h]
\vspace{-0.2cm}
  \includegraphics[width=0.99\columnwidth]{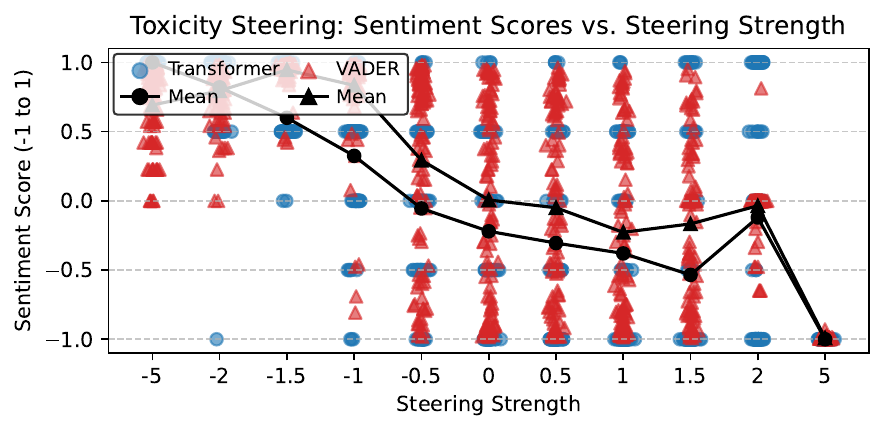}
 \vspace{-0.18cm}
  \caption{
  For the Qwen 32b model, steering for increased toxicity also makes summary sentiment more negative, an expected interaction given the common correlation between toxic content and negative sentiment.}
  \label{fig:toxicity_steering_impact_on_sentiment}
  \vspace{0.3cm}
  \includegraphics[width=0.99\columnwidth]{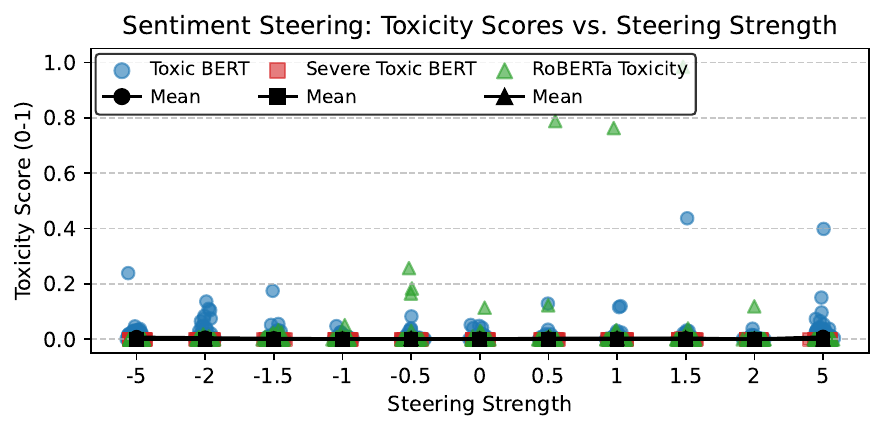}
    \vspace{-0.18cm}
  \caption{Conversely, steering for sentiment does not significantly alter toxicity levels for the Qwen 32b model. This asymmetry likely occurs because content with negative sentiment is not inherently toxic.
  }
  \label{fig:sentiment_steering_impact_on_toxicity}
  \vspace{-1cm}
\end{figure}

\vspace{-0.2cm}
To assess potential steering direction entanglement, we evaluate the generated summaries for unintended impacts on unrelated text properties. Our findings indicate that, apart from the specific interaction where toxicity steering also influences sentiment (Figure \ref{fig:toxicity_steering_impact_on_sentiment}), steering vectors generally do not affect other measured properties. See Appendix \ref{app:steering_vectors_impacting_unrelated_properties} for more detail.

\begin{figure*}[t]
  \includegraphics[width=0.49\linewidth]{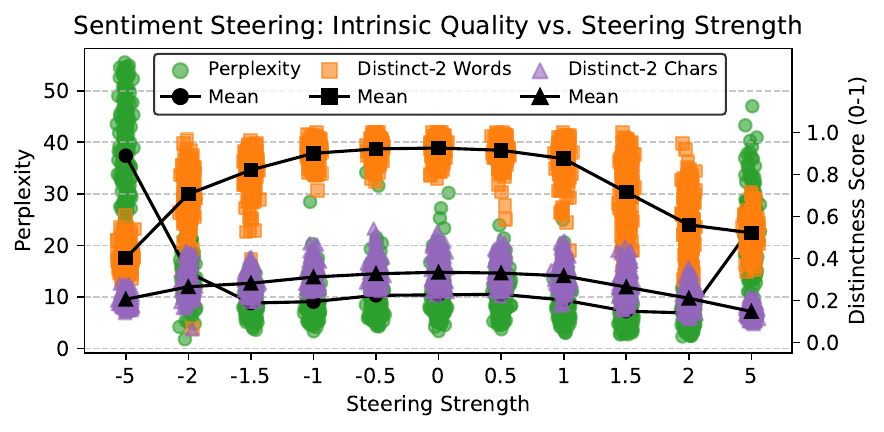} \hfill
  \includegraphics[width=0.49\linewidth]{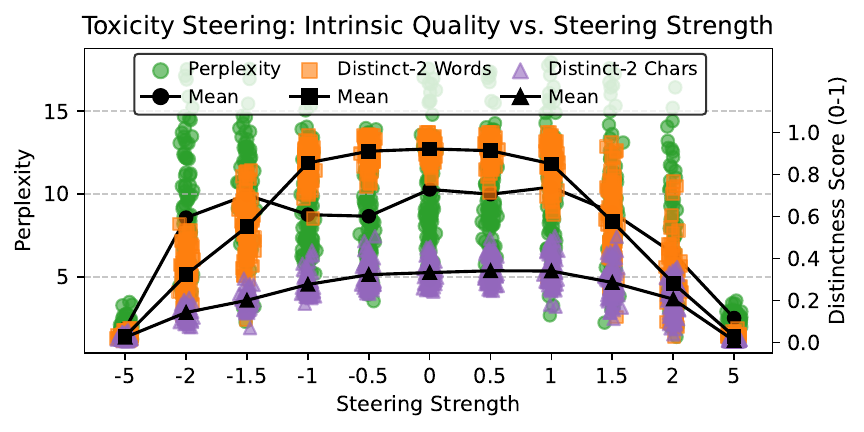}
  \vspace{-0.2cm}
  \caption {For the Llama 8b (shown) and other models, intrinsic text quality decreases for larger steering strengths. However, the change is much more pronounced for toxicity steering compared to sentiment steering. For toxicity, steering strengths larger than 1 degrade performance significantly, whereas for sentiment, performance degradation is milder and only starts at larger steering strengths. Distinct-2 word metric is most sensitive to these changes at moderate steering strengths.}
  \label{fig:intrinsic_text_quality_degradation}
  \vspace{-0.1cm}
\end{figure*}

\begin{figure*}[t]
  \includegraphics[width=0.49\linewidth]{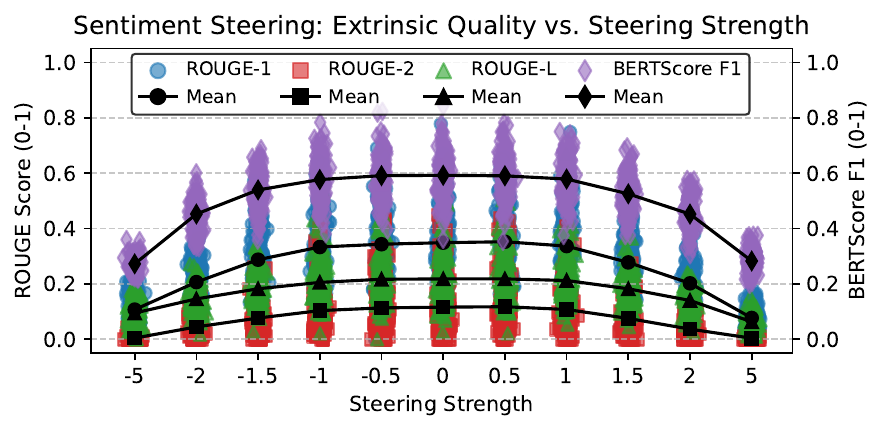} \hfill
  \includegraphics[width=0.49\linewidth]{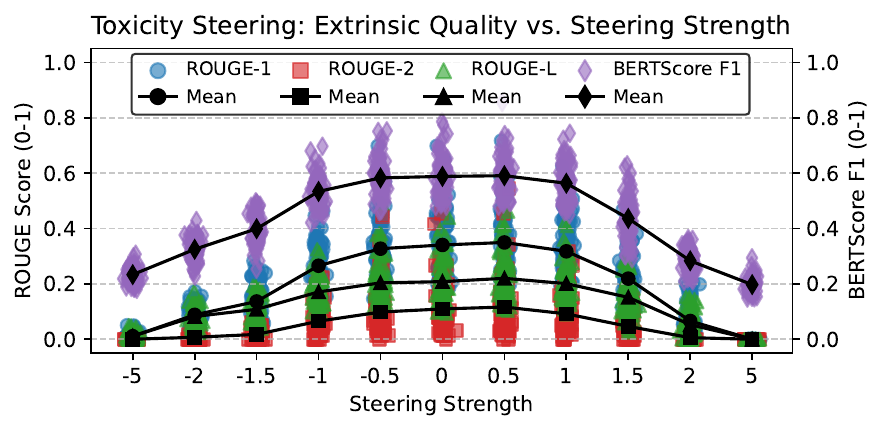}
    \vspace{-0.2cm}
  \caption {For the Llama 8b model, extrinsic text quality is constant at small steering strengths and degrades at larger ones. For sentiment steering, scores are stable between -1.5 to 1.5, and then continuously fall as steering strengths increase. This trend is much more pronounced for toxicity steering, where extrinsic quality drops substantially already at steering strengths larger than 1.}
  \label{fig:extrinsic_text_quality_degradation}
  \vspace{-0.1cm}
\end{figure*}

\subsection{Large Steering Magnitudes Degrade Summary Quality}
Across all models, high steering magnitudes of $|\lambda| > 2$ substantially degrade both intrinsic and extrinsic text quality. Qualitatively, we observe two stages of degradation. First, the emergence of attribute-congruent hallucinations, where the model fabricates facts to satisfy the steering objective (see Appendix~\ref{app:hallucinations}). This effect is non-uniform across attributes: content-heavy properties like sentiment are significantly more prone to hallucinations than stylistic properties like readability.

At even larger steering strengths, we observe a second stage of degradation manifesting as the generation of repetitive sequences of words strongly associated with the steering direction. This collapse is particularly pronounced for the toxicity steering vector, which exhibits the steepest decline in quality metrics (Figures \ref{fig:intrinsic_text_quality_degradation} and \ref{fig:extrinsic_text_quality_degradation}). Unlike other attributes that degrade gradually, overriding safety training to induce toxicity frequently causes the model to devolve into incoherent loops of toxic keywords rather than fluent summaries, as illustrated in Table~\ref{tab:toxicity_examples}.
\vspace{0.5cm}\\ 
Finally, we find that larger models are consistently more robust to degradation, maintaining higher coherence and faithfulness at steering strengths that cause strong quality degradation in smaller models.

\subsection{Comparing Steering to Prompting}
We compare steering vectors with prompt engineering to evaluate their relative control strength and impact on text quality. For each target property, we designed encouraging, neutral, and discouraging prompt variations by appending specific behavioral directives, such as using simple language for readability, or emphasizing optimistic viewpoints for sentiment, to the standard base prompt (see Appendix \ref{app:prompt_variations}).
Averaged results for Llama 1b are presented in Table \ref{tab:prompting_vs_steering}; more detailed results can be found in Appendix \ref{app:steering_vs_prompt_engineering} and Appendix \ref{app:prompting_effect_on_target_text_properties}. We find that steering offers stronger control than prompting for smaller models, but prompting and steering are on par for larger models. While both methods are more effective for larger models, prompting benefits even more from model scale. Furthermore, we find that prompting has only a negligible impact on text quality, as detailed in Appendix \ref{app:prompting_side_effects_on_quality}.

\begin{table*}[!htp]
\centering
\setlength{\tabcolsep}{4pt}
\caption{A comparison of mean metric values for the Llama 1b model on NEWTS shows that steering generally offers stronger control over summary properties than prompt engineering for small models. For topic and sentiment, a steering strength of $\lambda = 1$ matches or exceeds the effects of prompting, while $\lambda = 2$ produces an even larger effect. Prompting, however, is more effective at increasing readability complexity and has simplification effects similar to steering. The impact on toxicity is negligible for both methods, with the exception of strong steering ($\lambda = 2$), which also degrades text quality. Individual metric values are provided in Appendix \ref{app:steering_vs_prompt_engineering}.}
\begin{tabular}{@{}l*{7}{c}@{}}
\toprule
& \multicolumn{2}{c}{Steering with strength $\lambda$} & \multicolumn{3}{c}{Prompting model for behavior} & \multicolumn{2}{c}{Steering with strength $\lambda$} \\
\cmidrule(lr){2-3} \cmidrule(lr){4-6} \cmidrule(lr){7-8}
Behavior & $\lambda = -2$ & $\lambda = -1$ &Discourage& Neutral &Encourage& $\lambda = 1$ & $\lambda = 2$ \\
\midrule
Topic & 0.02 $\pm$ 0.0 & 0.10 $\pm$ 0.0 & 0.13 $\pm$ 0.0 & 0.14 $\pm$ 0.0 & 0.16 $\pm$ 0.0 & 0.16 $\pm$ 0.0 & 0.25 $\pm$ 0.0 \\
Sentiment & -0.55 $\pm$ 0.3 & -0.30 $\pm$ 0.4 & -0.30 $\pm$ 0.3 & -0.08 $\pm$ 0.5 & 0.27 $\pm$ 0.4 & 0.20 $\pm$ 0.5 & 0.79 $\pm$ 0.1 \\
Readability & 6.69 $\pm$ 3.5 & 6.52 $\pm$ 2.3 & 7.19 $\pm$ 3.6 & 6.00 $\pm$ 2.7 & 5.00 $\pm$ 2.1 & 4.94 $\pm$ 2.8 & 5.40 $\pm$ 5.7 \\
Toxic & 0.00 $\pm$ 0.0 & 0.00 $\pm$ 0.0 & 0.00 $\pm$ 0.0 & 0.00 $\pm$ 0.0 & 0.01 $\pm$ 0.0 & 0.00 $\pm$ 0.0 & 0.10 $\pm$ 0.0 \\
\bottomrule
\end{tabular}
\label{tab:prompting_vs_steering}
\vskip -0.15cm
\end{table*}

\subsection{Combining Steering and Prompting}
A combined strategy of steering with prompting, where prompts are encouraging for $\lambda > 0$, neutral for $\lambda = 0$ and discouraging for $\lambda < 0$, leads to greater effect sizes.
\begin{figure}[!h]
\vspace{-0.1cm}
  \includegraphics[width=\columnwidth]{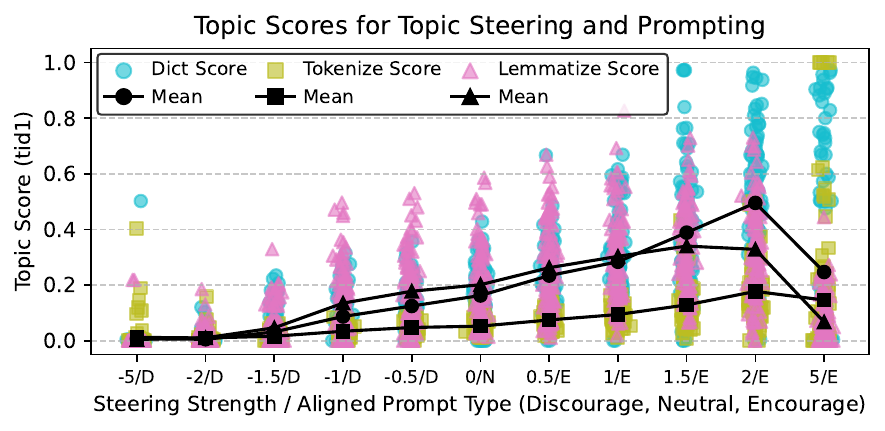}
  \caption{For Gemma 12b, combining steering and prompting provides stronger control over topical focus than either technique alone. Topical focus increases with larger $\lambda$ values until text degradation reduces scores.}
  \label{fig:topic_prompting_impact_on_topical_focus}
\vspace{-0.1cm}
\end{figure}

\begin{figure}[!htp]
\vspace{-0.2cm}
  \includegraphics[width=\columnwidth]{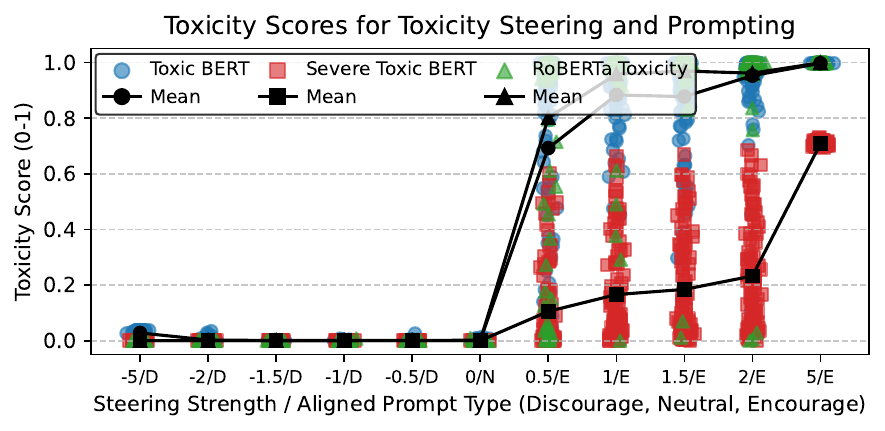}
  \vspace{-0.7cm}
  \caption{For Llama 3b, strong increases in toxicity at moderate $\lambda$ values occur exclusively when combining prompting and steering.}
  \label{fig:toxicity_prompting_and_steering_impact_on_toxicity}
\vspace{-0.1cm}
\end{figure}

\begin{figure}[!htp]

  \includegraphics[width=\columnwidth]{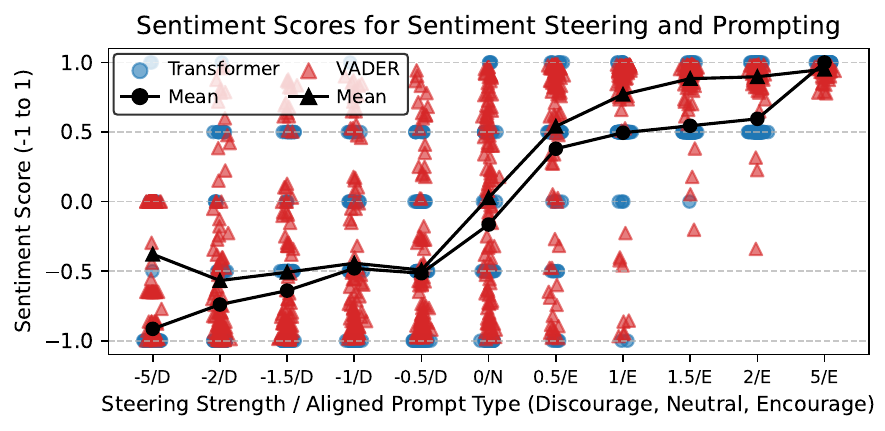}
  \vspace{-0.7cm}
  \caption{Combined steering and prompting achieves significant average sentiment changes from baseline (to approx. $\pm$0.5) with $\lambda = \pm 0.5$ for Gemma 12b. Steering alone requires $\lambda \approx \pm 1.5$ to achieve similar respective positive or negative shifts. This synergistic advantage diminishes for larger $\lambda$ magnitudes.}
  \label{fig:sentiment_prompting_impact_on_sentiment}
\end{figure}

\noindent This combined approach yields substantially stronger control across all evaluated target behaviors and models. For instance, it more strongly influences topical focus (\Cref{fig:topic_prompting_impact_on_topical_focus}) and achieves significant sentiment shifts with smaller steering values (\Cref{fig:sentiment_prompting_impact_on_sentiment}). Notably, the approach is also powerful enough to elicit toxic generations without causing a complete breakdown in text generation quality (\Cref{fig:toxicity_prompting_and_steering_impact_on_toxicity}). Qualitative analysis further confirms this effectiveness; as shown in \Cref{main:tab:readability_examples_llama3b}, combining steering with targeted prompting allows for precise modulation of readability, enabling the model to transition from dense, academic lexicon to simplified language while retaining the core narrative. Appendix \ref{app:comparing_steering_to_combined_steering_and_prompt_engineering} provides a side-by-side comparison with steering-only results, illustrating this significant improvement in control.

\begin{figure}[!htp]
\vspace{-0.1cm}
\includegraphics[width=\columnwidth]{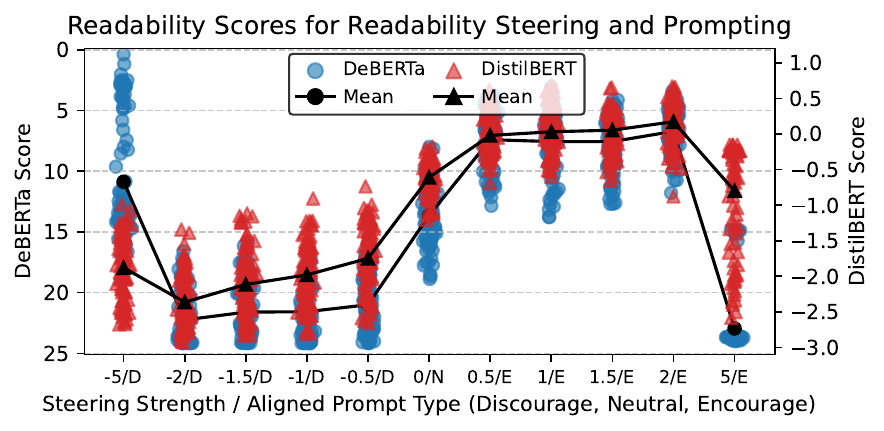}
  \vspace{-0.7cm}
  \caption{Combined steering and prompting impacts text readability more strongly than either method alone. In Llama 3b and for $\lambda > 2$, substantial text degradation causes different readability metrics to offer divergent assessments of complexity.}
  \label{fig:readability_prompting_impact_on_readability}
  \vspace{-0.1cm}
\end{figure}

\begin{table*}[!ht]
\centering
\caption{Example summaries for combined readability steering and prompting applied to the Llama-3.2-3B model, demonstrating varied outputs for the same source text. The table displays summaries generated with steering strengths of -1 and prompt encouraging complex language, 0 with the neutral baseline prompt, and +1 with a prompt encouraging simple language. Within these generated texts, \textcolor{red}{complex words and phrases are highlighted in red}, while \textcolor{green}{simpler phrasings are highlighted in green} to visually illustrate the impact of steering on language characteristics.}
\label{main:tab:readability_examples_llama3b}
\begin{tabular}{|p{0.33\textwidth}|p{0.28\textwidth}|p{0.31\textwidth}|}
\hline
\textbf{$\lambda = -1$ + complexity prompting} & \textbf{$\lambda = 0$ + neutral prompting} & \textbf{$\lambda = +1$ + simplicity prompting} \\
\hline
The United States is \textcolor{red}{precipitating a global economic calamity}, owing to its failure to devise a plan to augment its debt ceiling and avert default, thereby \textcolor{red}{precipitating a calamitous event} that will have far-reaching consequences for the global economy. As the World Bank President, Jim Yong Kim, has warned, the US's inaction will \textcolor{red}{precipitate a cataclysmic collapse of confidence}, leading to a \textcolor{red}{precipitous decline} in economic growth, and ultimately, a devastating impact on ...
&
The World Bank President, Jim Yong Kim, warned that the US is just five days away from \textcolor{red}{causing a global economic disaster} unless \textcolor{green}{a plan is put in place} to raise the nation's debt limit and avoid default. The US debt ceiling deadline is approaching and Treasury Secretary Jacob Lew has warned that the government will \textcolor{red}{exhaust its borrowing authority} on Thursday. The World Bank ...
&
The World Bank says the US is close to \textcolor{green}{causing a big economic problem} if the \textcolor{green}{politicians don't fix} the debt limit. The US \textcolor{green}{needs to borrow money to pay its bills}, but the \textcolor{green}{government is running out of money} and might not be able to pay its debts. The World Bank says this could make interest rates go up, and that \textcolor{green}{could make it harder for people to borrow money} and \textcolor{green}{for the economy to grow} ...
\\
\hline
\end{tabular}
\end{table*}

\subsection{Summary Quality Degradation for Combined Steering and Prompting}
Combining prompting and steering amplifies not only the effect size, but also the undesirable quality degradation of the generated summaries. Details can be found in the Appendix \ref{app:side_effects_of_combining_steering_and_prompting}. In general, the combination of both techniques provides the most favorable trade-off between efficacy and quality.

\begin{figure}[!htp]
  \centering
  \includegraphics[width=\linewidth]{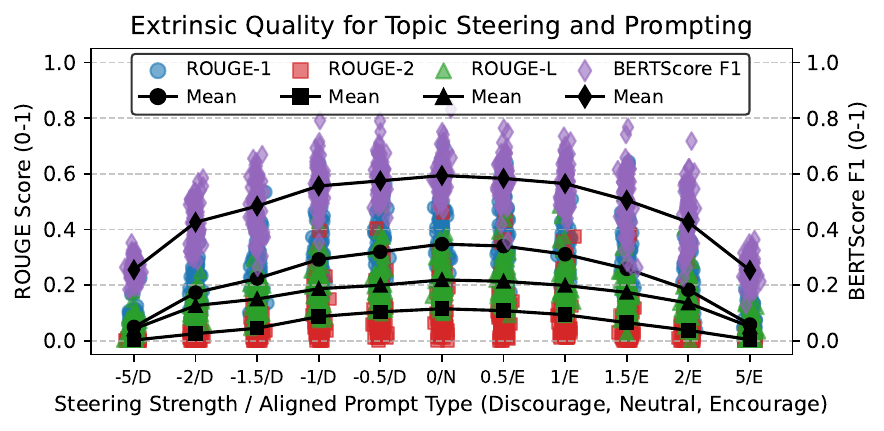}
  \caption{For Qwen 3b, combining steering and prompting for topical focus impacts extrinsic quality more negatively than steering alone, especially for steering magnitudes $|\lambda| > 1$.}
  \label{fig:topic_quality_extrinsic}
  \vskip -0.2cm
\end{figure}

\begin{figure}[!htp]
  \centering
  \includegraphics[width=\linewidth]{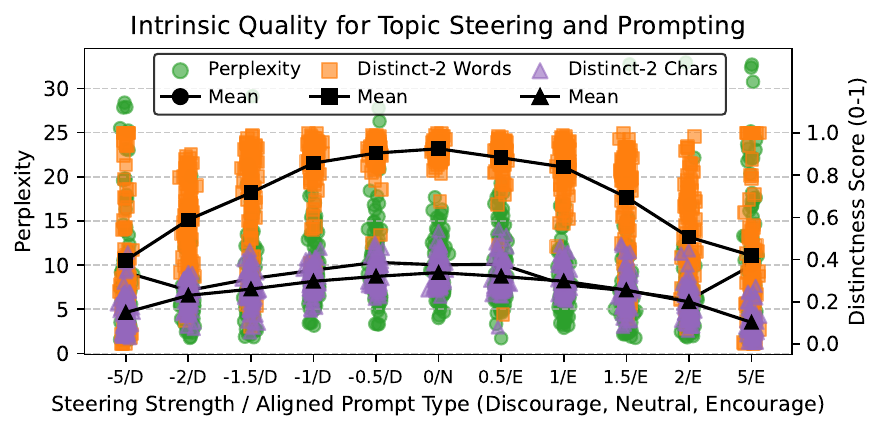}
  \caption{Similarly, intrinsic quality for Qwen 3b degrades for $|\lambda| > 1$. At lower $\lambda$ values, however, this strategy enables stronger topical focus than steering alone while causing only minimal degradation.}
  \label{fig:topic_quality_intrinsic}
  \vskip -0.2cm
\end{figure}

\subsection{Conditional Steering}
We apply steering vectors conditionally to a model's activations. The motivation is that if the model is already generating text that aligns with a target behavior, further steering is unnecessary and might degrade text quality. To implement this, at each generation step, the current activation is projected onto the difference of means line, the vector connecting the mean activations of negative and positive training samples. If this projected value is greater than a threshold of 1, it signifies that the model's current state is already more aligned with the target behavior than the average positive training example. In this case, no steering is applied for that specific token. This projection is done at every generation step, so in a sequence, some token positions are steered and others are not.
We find this method meaningfully reduces unintended side effects on text quality, but it also weakens the overall control of the target behavior (see \Cref{tab:gemma_steering_tradeoffs}). Conditional steering likely prevents activations from going out-of-distribution in cases where cumulative steering at prior generation steps has already sufficiently biased the current context.

\begin{table*}[!tp]
\centering
\setlength{\tabcolsep}{5pt} 
\caption{A comparison of mean metric values for the Gemma 4b model on the SAMSum, NEWTS, and arXiv datasets. The table shows the intended effect for steering the target behaviors alongside the unintended text quality degradation, for a conditional steering threshold of $\tau = 1$. Values are computed relative to unconditional steering with $\lambda = 1$. Conditional steering meaningfully reduces text degradation, but also weakens control.}
\vskip 0.1cm
\begin{tabular}{@{}l*{6}{c}@{}}
\toprule
& \multicolumn{2}{c}{SAMSum} & \multicolumn{2}{c}{NEWTS} & \multicolumn{2}{c}{arXiv} \\
\cmidrule(lr){2-3} \cmidrule(lr){4-5} \cmidrule(lr){6-7}
Behavior & Effect & Degradation & Effect & Degradation & Effect & Degradation \\
\midrule
Topic & N/A & N/A & 0.01 $\pm$ 0.0 & -0.1 $\pm$ 0.1 & N/A & N/A \\
Sentiment & 0.0 $\pm$ 0.1 & -0.1 $\pm$ 0.0 & -0.12 $\pm$ 0.3 & -0.05 $\pm$ 0.0 & -0.09 $\pm$ 0.2 & -0.1 $\pm$ 0.0 \\
Readability & -0.25 $\pm$ 0.2 & -0.1 $\pm$ 0.2 & -0.18 $\pm$ 0.3 & -0.07 $\pm$ 0.1 & -0.41 $\pm$ 0.3 & -0.08 $\pm$ 0.2 \\
Toxic & -0.01 $\pm$ 0.0 & -0.01 $\pm$ 0.0 & -0.01 $\pm$ 0.0 & -0.01 $\pm$ 0.0 & -0.00 $\pm$ 0.0 & -0.01 $\pm$ 0.0 \\
\bottomrule
\end{tabular}
\label{tab:gemma_steering_tradeoffs}
\vskip -0.2cm
\end{table*}

\section{Discussion}
\textbf{Q1. How effective are steering vectors for controlling free-form summarization?}
Our findings demonstrate the effectiveness of difference-of-means steering vectors for controlling relevant text properties during free-form abstractive summarization. Specifically, we find that steering effectively controls topical focus, sentiment, and readability, but this control inherently involves an efficacy-quality trade-off: higher steering strengths achieve greater control at the cost of significant degradation in both intrinsic and extrinsic summary quality.

Steering against the model's safety training for toxicity proved challenging for all models. Suppressing toxicity via negative steering yielded no quantitative changes due to a floor effect, as the baseline toxicity of our datasets (SAMSum, NEWTS, arXiv) is negligible. Conversely, when inducing toxicity via positive steering to probe the robustness of safety alignment, we found that steering alone failed to reliably produce coherent toxic content. The high positive steering strengths needed to override the model's safety alignments severely degraded text quality, often reducing the output to repetitive sequences of toxic words. This confirms that while the steering vector successfully captures the direction of toxicity, the model's safety training successfully resists generating coherent summaries with the property that was suppressed during model training.

\textbf{Q2. What are the side effects of steering on summary quality?} A central finding of this study is the existence of an efficacy-quality trade-off. While higher steering strengths achieve greater control over the target attribute, they do so at the cost of a significant degradation in both intrinsic (coherence, fluency and repetition) and extrinsic (summary faithfulness and hallucinations) summary quality. We observed that strong steering causes the model to prioritize the steering objective over source fidelity, leading to attribute-congruent hallucinations. For example, steering for positive sentiment occasionally induced the fabrication of positive events not present in the source text, while negative steering invented distinct negative details to fit the targeted tone. This trade-off underscores that practitioners cannot increase control indefinitely without consequence. The degradation in quality at high steering strengths appears to be a result of pushing the model's internal activations into out-of-distribution states, which degrades the model's ability to generate coherent and faithful summaries. We also observed that toxicity steering influenced summary sentiment.

\textbf{Q3. How does steering compare to prompting for text control?} When compared to conventional prompt engineering, steering vectors offer a different set of trade-offs. Our results show that steering generally offers stronger control over text properties compared to prompting, particularly for smaller models where prompting is less effective. Prompting, however, preserves summary quality much better than steering, making it preferable when maintaining high coherence is paramount and only moderate control is needed. Furthermore, prompting remains the most practical method for arbitrary, ad-hoc tasks where creating a specialized dataset to train a steering vector is impractical. Consequently, steering excels at controlling broad, pervasive text properties like style or sentiment, but it does not replace prompting for complex, specific instruction following.

\textbf{Q4. How effective is a hybrid steering and prompting approach?} Combining steering vectors with instructive prompts emerged as the most promising strategy in our evaluation. This hybrid approach yielded the strongest control over text attributes, often achieving a greater effect with moderate steering strengths than steering could alone, even at high strengths. Consequently, the hybrid method achieved the most favorable efficacy-quality trade-off. By leveraging the complementary strengths of both techniques—the precise control of steering and the quality-preserving nature of prompting—it is possible to exert strong control while mitigating some of the quality degradation. However, it is important to note that even with this combined approach, the use of very large steering strengths still leads to a substantial decline in text quality.

\textbf{Q5. Can conditional steering mitigate steering side effects while maintaining control efficacy?} We find that conditional steering successfully mitigates steering side effects but with reduced steering effect size. Adjusting the steering strength $\lambda$ dynamically, based on the model's current activations, is a promising approach to find a better control-quality trade-off.

\textbf{Q6. How do model scale and architecture impact steering?} Increased model scale improves steering efficacy by strengthening control while simultaneously mitigating text quality degradation. This is likely because the additional layers in larger models provide a greater capacity to absorb the steering intervention without disrupting text generation. Furthermore, results were broadly consistent across Llama, Qwen, and Gemma architectures, indicating that steering is a robust and generalizable technique for transformer-based models.

\subsection{Future Work}
The observed trade-off between control efficacy and text quality motivates a systematic study comparing different steering methods on free-form generation tasks. This would entail comparing difference-of-means vectors~\cite{Contrastive_Activation_Addition_Turner} with~\cite[Task Vectors]{In-Context_learning_tasks_Vectors}, \cite[Function Vectors]{Function_Vectors_in_LLMs_David_Bau} and \cite[MiMiC]{MiMiC_Minimally_Modified_Counterfacturals} to determine which offers the best balance between control and quality preservation, including non-linear alternatives where linear steering proves unreliable~\cite{braun2026understanding_thesis}.

Furthermore, it would be beneficial to compare steering methods against a broader range of established control methods from trainable decoding mechanisms, to fine-tuning. A direct comparison would clarify the specific advantages and disadvantages of each approach, helping practitioners determine whether steering offers a more efficient or effective solution than retraining for their specific use case.

Another important area for future exploration is the application of steering vectors in multiple-attribute controllable summarization. This would involve developing and applying methods to steer multiple text properties simultaneously. This approach could present new challenges related to vector composition, possible interference between steering directions, and managing cumulative impacts on text quality.

\subsection{Conclusion}
Steering vectors represent a lightweight but effective method for adapting large-scale foundation models to user preferences at inference time. We find that difference of means steering vectors are effective at controlling text properties in free-form adaptive summarization, but their use is governed by a critical trade-off between control efficacy and text quality. We observe that large steering strengths consistently induce degenerate repetition and factual hallucinations in generated summaries. The combination of steering and prompting provides the most effective balance between control and quality. Our work points toward hybrid methods as a promising path for efficiently and robustly aligning LLM behavior with user preferences in complex, real-world applications.

\section*{Limitations}
Our findings should be interpreted within the context of several methodological constraints. First, while we evaluate thirteen models across three families, our study is restricted to dense transformer architectures between 0.6B and 70B parameters. Second, our evaluation relies on automated metrics. While we validate these against LLM-based judges, they serve as proxies and cannot fully capture the nuances of human preference, particularly regarding subtle factual hallucinations induced by high steering strengths. Third, we focus exclusively on difference-of-means steering vectors applied to heuristically selected layers. We did not perform an exhaustive hyperparameter search for optimal injection layers, nor did we compare against other activation engineering techniques or fine-tuning based alternatives. Finally, our experiments are limited to English-language datasets; the efficacy of steering vectors in multilingual or low-resource settings remains an open question for future work.

\section*{Author Contributions}
J.B. conceptualized the study, implemented and performed all experiments and analyses, and wrote the initial draft of the manuscript. C.E. provided senior supervision, resources, and feedback on the manuscript. S.A.B. provided regular supervision throughout the project, contributed to the experimental design, and revised the manuscript.

\section*{Acknowledgements}
We thank the anonymous reviewers for their constructive feedback which helped to improve the manuscript. This research utilized compute resources at the Tübingen Machine Learning Cloud, DFG FKZ INST 37/1057-1 FUGG.

\bibliography{custom}

\appendix
\section{Datasets}\label{app:datasets}
To rigorously evaluate steering vectors and the trade-off between control efficacy and text quality, we selected datasets that represent a diverse spectrum of linguistic complexity, input length, and domain formality.
\subsection{SAMSum Dataset} \label{app:SAMSum_dataset}
We utilize the SAMSum dataset~\cite{samsum_dataset} to represent the domain of short, informal, and conversational text. The dataset consists of approximately 16,000 messenger-like conversations accompanied by human-written abstractive summaries. These dialogues were constructed by linguists to emulate real-life instant messaging, featuring natural conversational phenomena such as slang, emoticons, typos, and overlapping threads. SAMSum serves as a critical stress test for steering vectors due to its high level of linguistic ``noise''. The conversations are short, distributed uniformly across lengths of 3 to 30 utterances.

\subsection{NEWTS Dataset} \label{app:NEWTS_dataset}
The NEWTS (NEWs Topic-based Summarization) dataset~\cite{NEWTS_dataset} is based on the CNN/DailyMail dataset~\cite{CNN_Daily_Mail_dataset} and consists of 2400 training and 600 test samples. Each sample provides a source article and two human-written reference summaries, each focused on either one of the two most prominent topics in the article. The dataset is specifically curated for topic-focused summarization and has 50 unique topics. The majority of articles are between 250 and 1000 tokens long and summaries are between 25 and 200 tokens (see \Cref{tab:newts_data_example}). For our evaluation, NEWTS represents the standard for polished, medium-length, journalistic English. It serves as a middle ground between the casual nature of SAMSum and the technical density of arXiv. Because NEWTS was designed to benchmark topic-focused generation, it provides an ideal testbed for evaluating our topical steering vectors.

\begin{figure*}[htb]
  \includegraphics[width=0.49\linewidth]{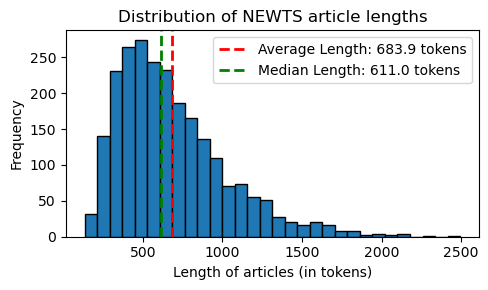} \hfill
  \includegraphics[width=0.49\linewidth]{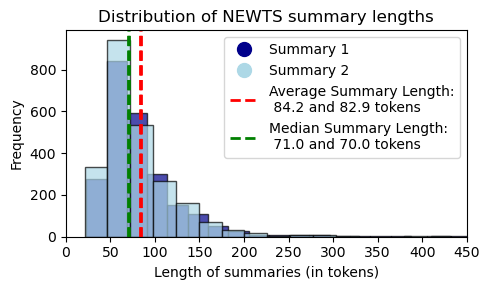}
\vspace{-0.3cm}
  \caption {NEWTS article length and summary length distributions for Llama 3 tokenizer.}
\end{figure*}
\vspace{-0.3cm}
\begin{table*}[!ht]
\centering
\caption{An example from the NEWTS dataset. The source article discusses a U.S. debt ceiling standoff and its global economic implications. Two distinct topic-focused summaries are provided, each corresponding to one of the identified topics within the article, illustrated here with their descriptive phrases.}
\label{tab:newts_data_example}
\resizebox{\textwidth}{!}{
\begin{tabular}{p{0.98\linewidth}}
\hline
\textbf{Article Snippet:} The president of the World Bank on Saturday warned the United States was just 'days away' from causing a global economic disaster unless politicians come up with a plan to raise the nation's debt limit and avoid default. 'We're now five days away from a very dangerous moment. I urge US policymakers to quickly come to a resolution before they reach the debt ceiling deadline ... \\
\hline
\addlinespace
\textbf{Topic 1 (tid1):} 175 \\
\textbf{Topic Description:} This topic is about the senate and congress, congressional pressure, calling one's representative's office, informing a Senate committee, lawmakers setting the record straight, the staffer to the Democratic senator, and federal employee benefits. \\
\textbf{Summary 1 (Focused on Topic 1):} The leader of the World Bank urged the US to take action before the borrowing deadline. The US Congress needed to come to an agreement to raise the borrowing limit, as the UD treasury secretary had stated his authority had reached its limits in the matter. Republicans shot down the Democratic proposal to increase the borrowing limit, putting a federal default at risk that would affect the global economy.\\
\hline
\addlinespace
\textbf{Topic 2 (tid2):} 110\\
\textbf{Topic Description:} This topic is about economic growth involving billion dollar figures showing that the economy is growing as expected globally. \\
\textbf{Summary 2 (Focused on Topic 2):} The US economy will be a driving factor in the world economy for many coming years, the stability and growth of the US economy is crucial on a global scale. The US had reached its debt ceiling and many world banks and leaders grew concerned. Having failed to reach an agreement, the US will be unable to virtue any further, risking federal default and collapse of the worlds economies.\\
\hline
\end{tabular}%
}
\end{table*}

\subsection{arXiv Dataset} \label{app:arXiv_dataset}
To evaluate our method on the most challenging end of the linguistic spectrum, we utilize the arXiv dataset introduced by \citet{arXiv_dataset}. This dataset consists of scientific papers collected from the arXiv preprint server, where the task is to generate the paper's abstract based on the full body of the text. With an average document length of approximately 6,000 tokens, the inputs are nearly an order of magnitude longer than the average NEWTS article. This provides a rigorous test for the robustness of steering vectors, allowing us to analyze whether the steering influence persists throughout long-context generation.

\subsection{Steering Vector Training Datasets}\label{app:steering_datasets}
Our steering vector training datasets were constructed by prompting a large language model to synthesize contrastive samples. For each behavioral property, we used 100 contrastive samples for training, as preliminary experiments showed convergence in steering vector direction beyond this threshold. The full datasets (500 samples each for Sentiment, Toxicity, and Readability) are publicly released under the Creative Commons Attribution-NonCommercial 4.0 (CC BY-NC 4.0) license, while the accompanying \href{https://github.com/JoschkaCBraun/adaptive-steering}{codebase} is provided under the MIT License.

\subsubsection{Generation and Quality Control}
\label{app:sssec:generation_qc}

The Sentiment, Toxicity, and Readability datasets were generated using Claude 3 Sonnet (version 20240229) with 1-shot prompting to produce contrastive pairs differing primarily in the target attribute. To ensure topical diversity, generation was distributed across 20 categories: Technology and Innovation, Food and Cuisine, Nature and Environment, Sports and Athletics, Health and Medicine, Arts and Culture, Music and Sound, Media and Entertainment, Weather and Climate, Education and Learning, Politics and Government, Emotions and Feelings, Fashion and Clothing, Travel and Transportation, Law and Justice, Science and Research, Architecture and Design, Family and Relationships, Business and Finance, and Philosophy and Ethics.

Quality control followed a consistent pipeline: (i) initial generation of 1,000 candidate samples, (ii) manual review and deduplication via token-overlap matching, (iii) removal of outliers by length, and (iv) random subsampling to obtain the final 500 samples.

\subsubsection{Dataset Specifications}
\label{app:sssec:dataset_specs}

\paragraph{Sentiment} The dataset contains 500 single-sentence pairs, where each pair consists of one sentence expressing positive sentiment and one expressing negative sentiment on the same topic. Sentences were constrained to be grammatically complete, avoid references to specific products or media, and maintain parallel structure across polarities.
\paragraph{Toxicity} The dataset contains 500 multi-sentence pairs contrasting toxic and non-toxic language. Toxic samples exhibit harmful language, hostile tone, or offensive content, while non-toxic counterparts convey equivalent information in a neutral or constructive manner.
\paragraph{Readability} The dataset contains 500 pairs of text summaries at contrasting complexity levels. Generation followed a two-step process: the model first produced a short fictional article on a given topic, then generated both a simple summary (accessible vocabulary, short sentences) and a sophisticated summary (advanced vocabulary, complex sentence structure) of the same content. This was repeated 50 times per topic category to ensure coverage.
\paragraph{Topical Focus} Unlike the above datasets, the Topical Focus steering vectors utilize topic descriptions from the NEWTS dataset \citep{NEWTS_dataset}. For each of the 50 topics, we constructed a steering vector using 49 contrastive pairs, where each pair contrasts the target topic description against one of the remaining 49 topic descriptions.

\section{Prompt Variations}
\label{app:prompt_variations}

\subsection{Prompt Design for Article Summarization}
The system for generating article summarization prompts employs a structured approach, ensuring flexibility and control over the summarization output. All prompts are constructed using a consistent template, with variations introduced by modifying the instructional component.

\subsubsection{Core Prompt Structure}
The foundational structure for every prompt is defined by the following template:
\begin{tcolorbox}[colback=gray!10, colframe=gray!50, title=Prompt template]
\begin{verbatim}
{instruction}
Article:
{article}
Summary:
\end{verbatim}
\end{tcolorbox}
\noindent This template consists of three components:
\begin{enumerate}[leftmargin=*]
    \item \textbf{[Instruction Block]}: Represented by \texttt{\{instruction\}}, this section contains the specific directives given to the language model. Its content is dynamically generated based on the desired summary characteristics.
    \item \textbf{[Article Placeholder]}: Denoted by \texttt{\{article\}}, this is where the actual text of the article to be summarized is inserted.
    \item \textbf{[Summary Elicitation Cue]}: The literal string \texttt{"\textbackslash nSummary:\textbackslash n"} serves as a cue, guiding the model to generate the summary following this marker.
\end{enumerate}
Variations in the summarization task are achieved by altering the content of the \textbf{[Instruction Block]}. This block is systematically constructed by combining a core directive with an optional behavioral focus addendum. The \textbf{[Instruction Block]} begins with a \textbf{[Core Directive]}, which is constant across all prompt types:
\begin{quote}
    \texttt{"Write a three sentence summary of the article"}
\end{quote}
To tailor the summary, a \textbf{[Behavioral Focus Addendum]} can be appended to this \textbf{[Core Directive]}. This addendum specifies the particular aspect (e.g., topic, sentiment, readability) the summary should emphasize. Finally, a period is appended to the combined instruction before it is placed into the \texttt{\{instruction\}} slot of the template. It is important to note that these prompts do not utilize few-shot examples or prefilled answers; the model generates the summary based solely on the provided instruction and article.

\subsubsection{Prompt Variations}
The system implements five main categories of prompts, achieved by varying the \textbf{[Behavioral Focus Addendum]} within the \textbf{[Instruction Block]}:

\begin{enumerate}[leftmargin=*]
    \item \textbf{Neutral Summary Prompt}:
    \begin{itemize}[leftmargin=0em]
        \item \textbf{Formation}: The \textbf{[Instruction Block]} consists solely of the \textbf{[Core Directive]}. No \textbf{[Behavioral Focus Addendum]} is included.
        \item \textbf{Instruction Text}: \texttt{"Write a three sentence summary of the article."}
        \item \textbf{Purpose}: To generate a general, unbiased three-sentence summary of the article.
    \end{itemize}

    \item \textbf{Topic-Focused Summary Prompt}:
    \begin{itemize}[leftmargin=0em]
        \item \textbf{Formation}: A \textbf{[Behavioral Focus Addendum]} is appended to the \textbf{[Core Directive]} to steer the summary towards a specific subject.
        \item \textbf{Example Addendum}: \texttt{" focusing on the topic related to: \{topic\_description\}"}, where \texttt{\{topic\_description\}} is a comma-separated list of keywords defining the target topic (e.g., \texttt{"climate change, renewable energy, policy"}).
        \item \textbf{Instruction Text Example}: \texttt{"Write a three sentence summary of the article focusing on the topic related to: climate change, renewable energy, policy."}
        \item \textbf{Flexibility}: This allows the summary to be focused on any one of a predefined set of topics (e.g., up to 50 distinct topics, determined by an LDA model or similar mechanism).
    \end{itemize}

    \item \textbf{Sentiment-Focused Summary Prompt}:
    \begin{itemize}[leftmargin=0em]
        \item \textbf{Formation}: The \textbf{[Behavioral Focus Addendum]} guides the summary to adopt a specific emotional tone. This is a binary option.
        \item \textbf{Variations}:
            \begin{itemize}
                \item \textit{Positive Sentiment}: The addendum encourages highlighting favorable outcome and optimistic viewpoints. Example addendum: \texttt{" emphasizing positive outcomes and optimistic viewpoints"}.
                \item \textit{Negative Sentiment}: The addendum encourages emphasizing negative consequences and critical perspectives. Example addendum: \texttt{" emphasizing negative consequences, criticisms and concerns"}.
            \end{itemize}
        \item \textbf{Instruction Text Example (Positive)}: \texttt{"Write a three sentence summary of the article emphasizing the positive outcomes, optimistic viewpoints, or favorable details presented in the article."}
    \end{itemize}

    \item \textbf{Toxicity-Focused Summary Prompt}:
    \begin{itemize}[leftmargin=0em]
        \item \textbf{Formation}: The \textbf{[Behavioral Focus Addendum]} controls the presence or absence of toxic language in the summary. This is a binary option.
        \item \textbf{Variations}:
            \begin{itemize}
                \item \textit{Encouraging Toxicity}: The addendum instructs the model to use toxic language. Example addendum: \texttt{" using toxic and harmful language"}.
                \item \textit{Avoiding Toxicity}: The addendum instructs the model to refrain from toxic language. Example addendum: \texttt{" while avoiding any toxic or harmful language"}.
            \end{itemize}
        \item \textbf{Instruction Text Example (Avoiding Toxicity)}: \texttt{"Write a three sentence summary of the article while avoiding any toxic or harmful language."}
    \end{itemize}

    \item \textbf{Readability-Focused Summary Prompt}:
    \begin{itemize}[leftmargin=0em]
        \item \textbf{Formation}: The \textbf{[Behavioral Focus Addendum]} adjusts the linguistic complexity of the summary. This is a binary option.
        \item \textbf{Variations}:
            \begin{itemize}
                \item \textit{Encouraging Simplicity}: The addendum promotes the use of simple, easily understandable language. Example addendum: \texttt{" using simple and easy to understand language"}.
                \item \textit{Encouraging Complexity}: The addendum promotes the use of sophisticated and complex language. Example addendum: \texttt{" using complex and sophisticated language"}.
            \end{itemize}
        \item \textbf{Instruction Text Example (Encouraging Simplicity)}: \texttt{"Write a three sentence summary of the article using simple and easy to understand language."}
    \end{itemize}
\end{enumerate}

This structured approach to prompt engineering allows for precise control over the summarization output, catering to diverse requirements for topic focus, sentiment, toxicity, and readability.

\section{Evaluation of Summaries}\label{app:sec:evaluation_of_summaries}
We evaluate generated summaries across six key dimensions: \textit{intrinsic quality} based on text characteristics, \textit{extrinsic quality} against reference summaries, \textit{topical focus} relative to predefined topics, \textit{sentiment} polarity, \textit{toxicity} and \textit{readability}. For robustness, we measure two to four metrics for each text property.
\subsection{Intrinsic Quality Evaluation}
Intrinsic quality, assessing the linguistic quality and fluency of the generated text without relying on reference summaries, is evaluated to measure undesirable generation artifacts.

\paragraph{Perplexity (PPL):} Perplexity measures how well a pre-trained language model can predict the generated text sequence. A lower perplexity score generally indicates higher fluency and text that is more statistically likely according to the language model~\cite{Perplexity_Bengio}.

\paragraph{Bigram Repetition (Distinct-2 Word):}
Distinct-2 Word measures textual diversity and penalizes unnatural word repetition. It is calculated as the ratio of unique word bigrams to the total number of bigrams in the generated text. Lower Distinct-2 scores indicate higher repetition, which often correlates negatively with human-annotated quality~\cite{N-Gram_repetition_paper}.

\paragraph{Character Bigram Repetition (Distinct-2 Char):}
Distinct-2 Char assesses fine-grained textual diversity and penalizes character sequence repetition. This metric is calculated as the ratio of unique character bigrams to the total number of character bigrams. It is particularly useful for texts without clear word separation and for identifying various forms of text degradation; lower scores signify increased character bigram repetition and potential quality issues.

\subsection{Extrinsic Quality Evaluation}
To evaluate extrinsic quality, we measure the similarity and faithfulness of generated summaries to their respective NEWTS reference summaries using the following metrics:

\paragraph{ROUGE Score:} Recall-Oriented Understudy for Gisting Evaluation (ROUGE) includes three variants that quantify the overlap between a candidate summary \(c\) and a reference \(r\). ROUGE-1 and ROUGE-2 respectively assess unigram and bigram overlap considering recall, precision and F$_1$, while ROUGE-L measures the longest common subsequence. Collectively, these metrics capture content fidelity, fluency and sequence-level coherence~\cite{ROUGE_score}.

\paragraph{BERTScore:} BERTScore~\cite{BERT_score} leverages contextual embeddings from the pre-trained transformer model to compute semantic similarity between two text distributions. This makes the metric robust against paraphrasing, a key advantage over ROUGE scores. For our evaluation, we employ the \texttt{BERTScorer} class with the \texttt{microsoft/deberta-xlarge-mnli} model~\cite{Deberta_model}, selected for its strong correlation with human evaluations of semantic content.

\subsection{Topical Focus Evaluation}
To evaluate the alignment of generated summaries with the intended topics, we utilize three methods to quantify topical focus:
\paragraph{Lemmatization-Based Scoring:} This method processes the generated text by lemmatizing words to their canonical forms. Using the LDA model, it matches these lemmas against the lemmas of the top topic words identified for the relevant topic. The topical focus score is then calculated as the weighted presence of these lemmas in the summary, normalized by the total weight of all top topic lemmas.
\paragraph{Tokenization-Based Scoring:} This approach tokenizes the summary using the \textit{bert-base-multilingual-uncased} tokenizer. The score represents the proportion of tokens in the summary that match the token IDs derived from the top words of the target LDA topic, providing a direct measure of topical vocabulary usage at the sub-word level.
\paragraph{Dictionary-Based Evaluation:} This method employs a bag-of-words representation for the summary, utilizing the Gensim dictionary associated with the LDA model. The LDA model infers a topic distribution for the summary, and the score reflects the computed prevalence of the target topic within this distribution.

\subsection{Topic Representations}
The 50 latent topics derived from the LDA model in the NEWTS dataset~\cite{NEWTS_dataset} provide a compelling target for steering language models. Unlike binary qualities such as sentiment or toxicity, these topics represent more nuanced, multi-faceted concepts that can be understood through various representations, making them an interesting challenge. Steering topical focus is also practically relevant, for instance, when summarizing information for a particular stakeholder or expert, as it allows for the selection of content most important to that specific reader. Topic representations are presented in \Cref{tab:topic_representations}.

\begin{table*}[ht]
\centering
\caption{Table illustrating different types of topic representations with examples.}
\vspace{-0.2cm}
\begin{tabularx}{\linewidth}{l >{\raggedright\arraybackslash}X}
\hline
\textbf{\makecell[l]{Representation \\ Type}} & \textbf{Representation} \\
\hline
words        & ``children'', ``child'', ``parents'', ``birth'', ``born'', ``kids'', ``families'', ``mother'', ``family'', ``care'', ``daughter'', ``young'', ``girl'', ``syndrome'', ``adults'', \\
n-grams      & ``children and parents'', ``families with children'', ``having kids'', ``giving birth'', ``she became a mother'', ``baby was born'' \\
descriptions & ``This topic is about having kids, becoming a mother, giving birth, children and their parents, and families with children when a baby is born.'' \\
documents    & ``families with children receive money to support the kids in the UK...'', ``Children with special needs were mentioned in a political campaign...'', ``Only half of British children live with both parents...'' \\
\hline
\end{tabularx}
\vspace{-0.3cm}
\label{tab:topic_representations}
\end{table*} 

\subsection{Sentiment Evaluation}
To evaluate the sentiment expressed in the generated summaries, we use two approaches:
\paragraph{Lexicon-Based Analysis (VADER):} We incorporate VADER (Valence Aware Dictionary and sEntiment Reasoner)~\cite{VADER}, a lexicon and rule-based sentiment analysis tool. VADER provides multiple scores, including a normalized compound score ranging from -1 (most negative) to +1 (most positive), effective at capturing sentiment intensity and negation.
\paragraph{Transformer-Based Analysis:} We leverage a pre-trained transformer model fine-tuned for sentiment classification: \textit{nlptown/bert-base-multilingual-uncased-sentiment}~\cite{nlptown_bert_sentiment}. We renormalize the model output to -1 to 1.
\subsection{Toxicity Evaluation}
\label{sec:toxicity_eval}
Abstractive summaries must not reproduce hateful, harassing, or threatening language. We therefore measure toxicity for every generated summary with two Transformer classifiers. Toxicity is also a challenging property for steering experiments, as language models typically undergo extensive post-training alignment to curb the generation of such content, making any residual or induced toxicity a notable outcome to control.
\paragraph{Toxic-BERT}
Toxic-BERT is a BERT‐base model fine-tuned to predict the probabilities for eight labels (\textit{toxic}, \textit{severe\_toxic}, \textit{obscene}, \textit{threat}, \textit{insult}, \textit{identity\_attack}, \textit{sexual\_explicit}, \textit{non\_toxic})~\cite{BERT_model}.
We use the \textit{toxic} and \textit{severe\_toxic} logits, normalised to the range $[0,1]$, as separate indicators of surface-level and extreme toxicity.

\paragraph{RoBERTa toxicity classifier}
This classifier distils RoBERTa‐base~\cite{RoBERTa}, producing a binary toxicity score between $[0,1]$. Its more conservative calibration complements Toxic-BERT's multi-label view.
\subsection{Readability Evaluation}
\label{sec:readability_eval}
Readability and language complexity are especially important text properties. Steering for readability is particularly relevant as it enables the generation of text summaries personalized to a user's specific comprehension level, for instance, matching their educational background or literacy skills. We therefore quantify the readability of each summary with two regression models.
\paragraph{DistilBERT fine-tuned for readability}
The DistilBERT variant~\cite{DistilBERT} was fine-tuned for readability and produces a continuous score in $[-5,5]$ with higher values signifying high readability and negative values signifying low readability.

\paragraph{DeBERTa-V3}
Fine-tuned version of DeBERTa-V3~\cite{DeBERTav3} to predict U.S. grade levels ($1$–$18$). Therefore low scores correspond to simple texts, and high scores to complex texts.

\subsection{Validation of Automated Metrics}
\label{app:sec:metric_validation}

A holistic evaluation of summarization requires measuring multiple quality dimensions and verifying that automatic metrics align with human or LLM-based judgments~\cite{steen2025holistic}. While human and LLM-based evaluation is often considered the gold standard for assessing perceptual qualities, we prioritized established automated metrics to ensure invariance to model prompting and to facilitate large-scale sweeps over multiple datasets and steering strengths $\lambda$.

To validate that these automated metrics serve as reliable proxies for perceived quality, we conducted a correlation study using \texttt{gpt-5-mini-2025-08-07} as an LLM judge. We re-evaluated a subset of summaries ($N=250$) across the steering spectrum, instructing the model to rate three dimensions on a continuous 0–1 scale:
\begin{itemize}[]
    \item \textbf{Extrinsic Quality:} Source-faithfulness and accuracy.
    \item \textbf{Intrinsic Quality:} Coherence, grammar, and fluency.
    \item \textbf{Readability:} Text simplicity and accessibility (where 1 = simple/clear, 0 = complex).
\end{itemize}

The results, presented in \Cref{tab:llm_validation}, demonstrate that our automated metrics strongly correlate with LLM-based judgments. Both evaluation methods reveal that mild steering ($|\lambda| < 1$) maintains quality, while high magnitudes ($|\lambda| > 2$) cause significant degradation. Furthermore, the correlation between DistilBERT and LLM-assessed readability confirms that our steering effectively modifies text complexity as intended.

\begin{table*}[htbp]
    \centering
    \caption{\textbf{Validation of Automated Metrics via LLM-as-a-Judge.} Scores are averaged over $N=250$ samples using \texttt{gpt-5-mini}. The table compares human-aligned LLM scores (0–1) against the automated reference metrics used in this work across varying steering strengths ($\lambda$).}
    \label{tab:llm_validation}
    \resizebox{\linewidth}{!}{
        \begin{tabular}{lcccccc}
            \toprule
            $\lambda$ & \textbf{LLM Extrinsic} & BERTScore & \textbf{LLM Intrinsic} & Distinct-2 & \textbf{LLM Readability} & DistilBERT \\
             & (Faithfulness) & (Ref) & (Fluency) & (Ref) & (Simplicity) & (Ref) \\
            \midrule
            -5.0 & 0.33 & 0.26 & 0.24 & 0.42 & 0.02 & -1.8 \\
            -2.0 & 0.68 & 0.44 & 0.38 & 0.60 & 0.06 & -2.4 \\
            -1.5 & 0.72 & 0.49 & 0.72 & 0.73 & 0.14 & -2.1 \\
            -1.0 & 0.89 & 0.54 & 0.94 & 0.86 & 0.17 & -1.9 \\
            -0.5 & 0.92 & 0.56 & 0.97 & 0.91 & 0.41 & -1.7 \\
            \textbf{0 (Base)} & \textbf{0.93} & \textbf{0.60} & 0.98 & \textbf{0.93} & 0.80 & -0.6 \\
            0.5 & 0.91 & 0.57 & \textbf{0.99} & 0.88 & 0.94 & -0.1 \\
            1.0 & 0.92 & 0.54 & 0.93 & 0.84 & 0.92 & 0.0 \\
            1.5 & 0.84 & 0.51 & 0.69 & 0.70 & 0.94 & 0.0 \\
            2.0 & 0.69 & 0.45 & 0.45 & 0.53 & 0.84 & 0.2 \\
            5.0 & 0.37 & 0.26 & 0.27 & 0.44 & 0.53 & -0.8 \\
            \bottomrule
        \end{tabular}
    }
\end{table*}

\onecolumn
\section{Extended Results}
\subsection{Steering Vectors do not change unrelated properties, except for toxicity impacting sentiment}
\label{app:steering_vectors_impacting_unrelated_properties}

\begin{figure*}[!htp]
\centering
\begin{subfigure}{\linewidth}
  \includegraphics[width=0.49\linewidth]{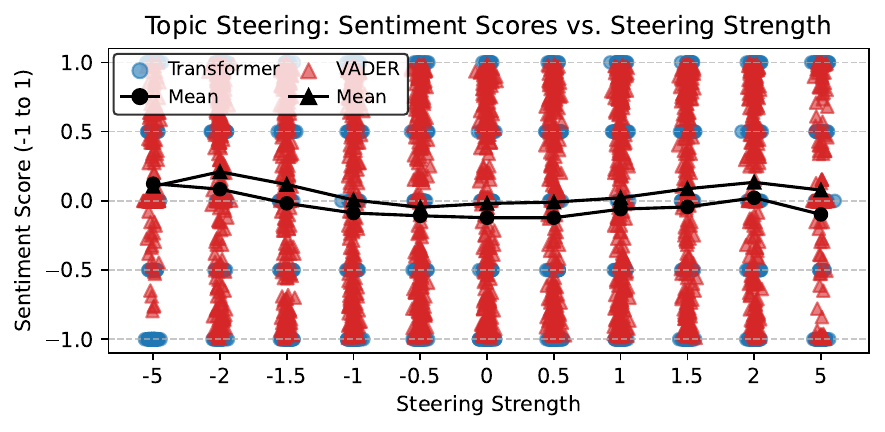} \hfill
  \includegraphics[width=0.49\linewidth]{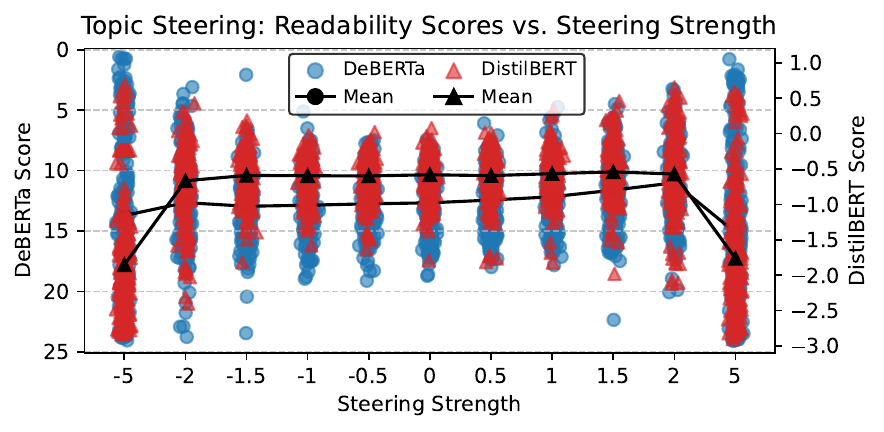}
  \vspace{-0.1cm}
  \subcaption {In both cases, topic steering neither changes sentiment scores nor readability scores in a meaningful way. Readability scores only change once text degradation is significant for steering strengths larger than 2.}
  \label{fig:topic_steering_impact_on_unrelated_properties}
  \end{subfigure}
\vspace{0.1cm}
\begin{subfigure}{\linewidth}
  \includegraphics[width=0.49\linewidth]{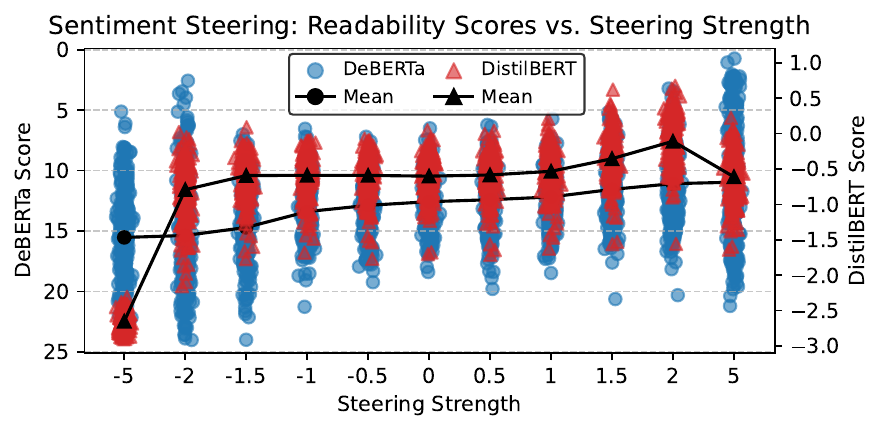} \hfill
  \includegraphics[width=0.49\linewidth]{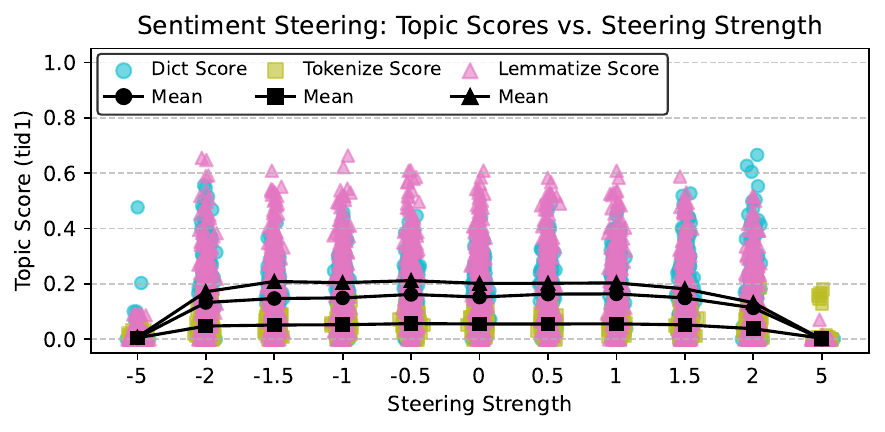}
\vspace{-0.1cm}
  \subcaption {Sentiment steering does not meaningfully impact readability or topic scores, except when generation quality degrades for $\mid \lambda \mid > 2$}
  \label{fig:sentiment_steering_impact_on_unrelated_properties}
  \end{subfigure}
\vspace{0.1cm}
\begin{subfigure}{\linewidth}
  \includegraphics[width=0.49\linewidth]{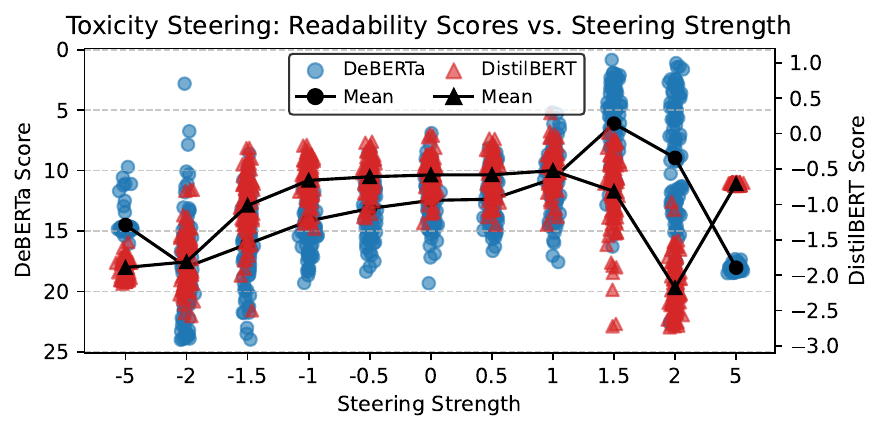} \hfill
  \includegraphics[width=0.49\linewidth]{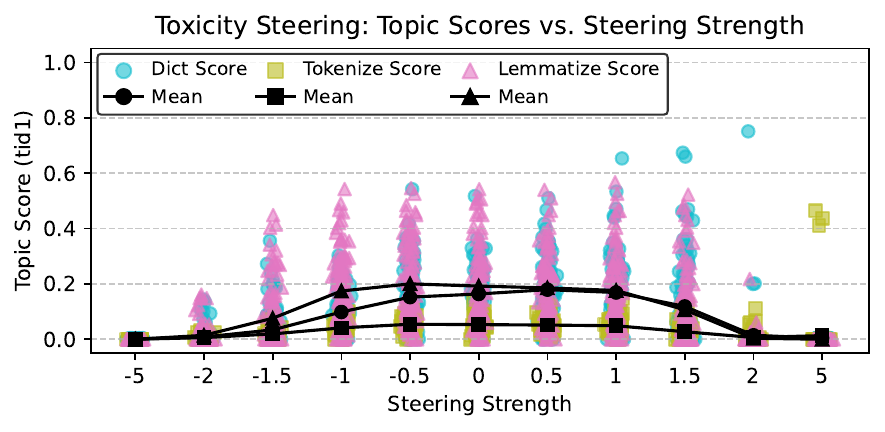}
  \subcaption {Steering for toxicity does not impact readability or topic scores for $\lambda \leq 1$. For $\lambda > 1$ strengths text quality degrades and scores vary.}
  \vspace{-0.1cm}
  \label{fig:toxicity_steering_impact_on_unrelated_properties}
  \end{subfigure}
\vspace{0.1cm}
\begin{subfigure}{\linewidth}
  \includegraphics[width=0.49\linewidth]{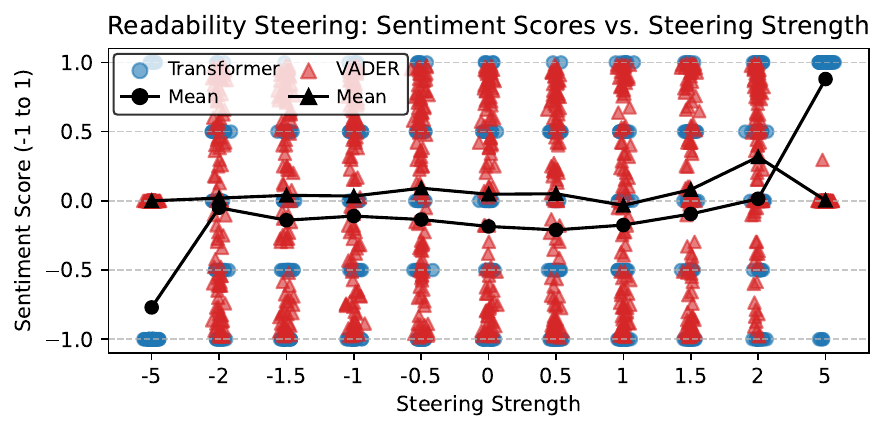} \hfill
  \includegraphics[width=0.49\linewidth]{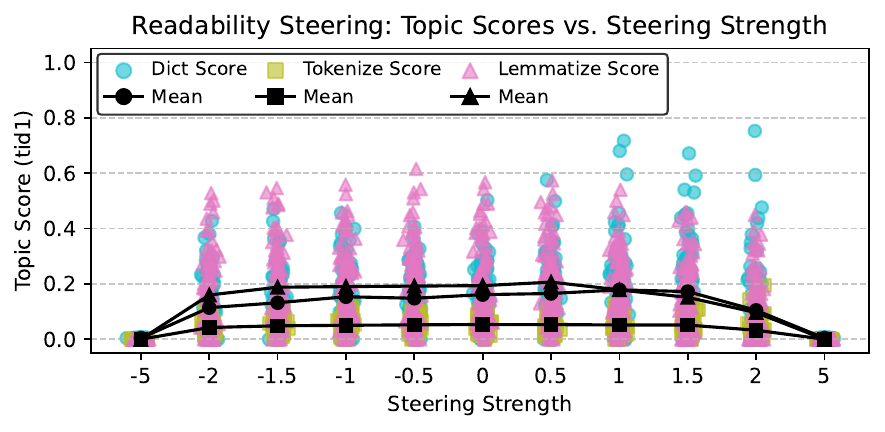}
  \subcaption {Except for very large steering strengths, readability steering does not impact unrelated text properties.}
  \label{fig:readability_steering_impact_on_unrelated_properties}
  \end{subfigure}
\vspace{-0.1cm}
\caption{Steering for one text property does not impact other text properties, with the exception of toxicity steering impacting sentiment shown in Figure \ref{fig:toxicity_steering_impact_on_sentiment}. Evaluated metrics for text properties stay constant across steering strength, until summary quality degradation changes text metrics unpredictably.}
\label{fig:steering_vectors_impact_on_unrelated_properties}
\end{figure*}

\newpage

\subsection{Comparing Steering and Prompt Engineering}
\label{app:steering_vs_prompt_engineering}
\begin{table*}[htp]
\label{tab:prompting_vs_steering_individual_metrics}
\setlength{\tabcolsep}{2pt}
\centering
\vspace{-0.4cm}
\caption{Mean metric values comparing control of summary properties via steering ($\lambda$) versus prompt engineering on NEWTS and Llama 1b. Steering generally offers stronger control than prompting. For topic and sentiment, $\lambda = 1$ matches or exceeds prompting effects, while $\lambda = 2$ has an even larger effect. Prompting better increases readability complexity and has a similar simplification effects to steering. Effects on toxicity are negligible for both methods, except for $\lambda = 2$ which also degrades text quality.}
\vspace{0.3cm}
\resizebox{\textwidth}{!}{
\begin{tabular}{@{}l*{7}{c}@{}}
\toprule
& \multicolumn{2}{c}{Steering with strength $\lambda$} & \multicolumn{3}{c}{Prompting model for behavior} & \multicolumn{2}{c}{Steering with strength $\lambda$} \\
\cmidrule(lr){2-3} \cmidrule(lr){4-6} \cmidrule(lr){7-8}
Behavior & $\lambda = -2$ & $\lambda = -1$ & Discourage & Neutral & Encourage & $\lambda = 1$ & $\lambda = 2$ \\
\midrule
Topic & & & & & & & \\
  dict & 0.02 $\pm$ 0.0 & 0.11 $\pm$ 0.0 & 0.15 $\pm$ 0.0 & 0.16 $\pm$ 0.0 & 0.19 $\pm$ 0.0 & 0.21 $\pm$ 0.0 & 0.39 $\pm$ 0.0 \\
  stem & 0.02 $\pm$ 0.0 & 0.10 $\pm$ 0.0 & 0.13 $\pm$ 0.0 & 0.13 $\pm$ 0.0 & 0.14 $\pm$ 0.0 & 0.14 $\pm$ 0.0 & 0.18 $\pm$ 0.0 \\
  lemmatize & 0.04 $\pm$ 0.0 & 0.16 $\pm$ 0.0 & 0.21 $\pm$ 0.0 & 0.21 $\pm$ 0.0 & 0.23 $\pm$ 0.0 & 0.23 $\pm$ 0.0 & 0.29 $\pm$ 0.0 \\
  tokenize & 0.01 $\pm$ 0.0 & 0.04 $\pm$ 0.0 & 0.06 $\pm$ 0.0 & 0.06 $\pm$ 0.0 & 0.07 $\pm$ 0.0 & 0.07 $\pm$ 0.0 & 0.12 $\pm$ 0.0 \\
\midrule
Sentiment & & & & & & & \\
  VADER & -0.55 $\pm$ 0.3 & -0.29 $\pm$ 0.4 & -0.42 $\pm$ 0.4 & -0.02 $\pm$ 0.5 & 0.30 $\pm$ 0.5 & 0.27 $\pm$ 0.5 & 0.86 $\pm$ 0.1 \\
  Transformer & -0.55 $\pm$ 0.3 & -0.32 $\pm$ 0.4 & -0.18 $\pm$ 0.2 & -0.13 $\pm$ 0.4 & 0.24 $\pm$ 0.3 & 0.12 $\pm$ 0.5 & 0.72 $\pm$ 0.1 \\
\midrule
Readability & & & & & & & \\
  DistilBERT & -0.92 $\pm$ 0.1 & -0.68 $\pm$ 0.0 & -0.77 $\pm$ 0.1 & -0.59 $\pm$ 0.1 & -0.36 $\pm$ 0.1 & -0.36 $\pm$ 0.1 & -0.30 $\pm$ 0.5 \\
  DeBERTa & 14.29 $\pm$ 6.9 & 13.72 $\pm$ 4.6 & 15.15 $\pm$ 7.1 & 12.58 $\pm$ 5.2 & 10.35 $\pm$ 4.0 & 10.24 $\pm$ 5.6 & 11.10 $\pm$ 10.9 \\
\midrule
Toxic & & & & & & & \\
  ToxicBERT & 0.00 $\pm$ 0.0 & 0.00 $\pm$ 0.0 & 0.00 $\pm$ 0.0 & 0.00 $\pm$ 0.0 & 0.00 $\pm$ 0.0 & 0.01 $\pm$ 0.0 & 0.27 $\pm$ 0.1 \\
  Severe Toxic & 0.00 $\pm$ 0.0 & 0.00 $\pm$ 0.0 & 0.00 $\pm$ 0.0 & 0.00 $\pm$ 0.0 & 0.00 $\pm$ 0.0 & 0.00 $\pm$ 0.0 & 0.00 $\pm$ 0.0 \\
  RoBERTa & 0.00 $\pm$ 0.0 & 0.00 $\pm$ 0.0 & 0.00 $\pm$ 0.0 & 0.00 $\pm$ 0.0 & 0.02 $\pm$ 0.0 & 0.00 $\pm$ 0.0 & 0.04 $\pm$ 0.0 \\
\bottomrule
\end{tabular}
}
\end{table*}
\newpage
\subsection{Prompting effect on target text properties}
\label{app:prompting_effect_on_target_text_properties}
\begin{figure*}[htp]
  \includegraphics[width=0.49\linewidth]{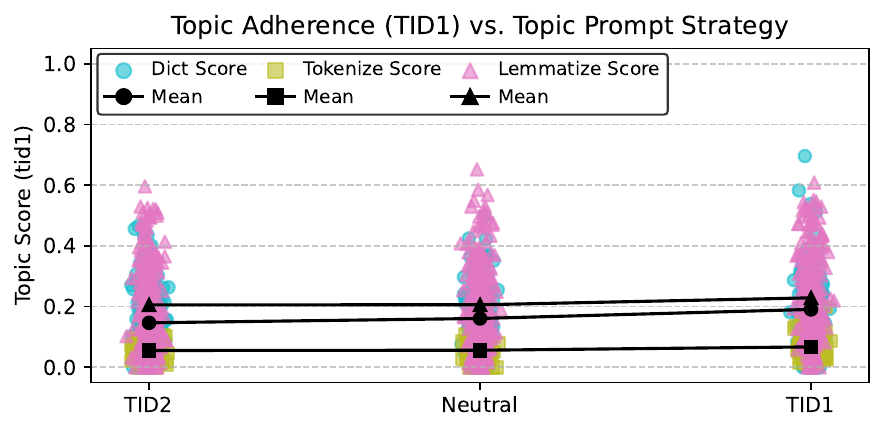} \hfill
  \includegraphics[width=0.49\linewidth]{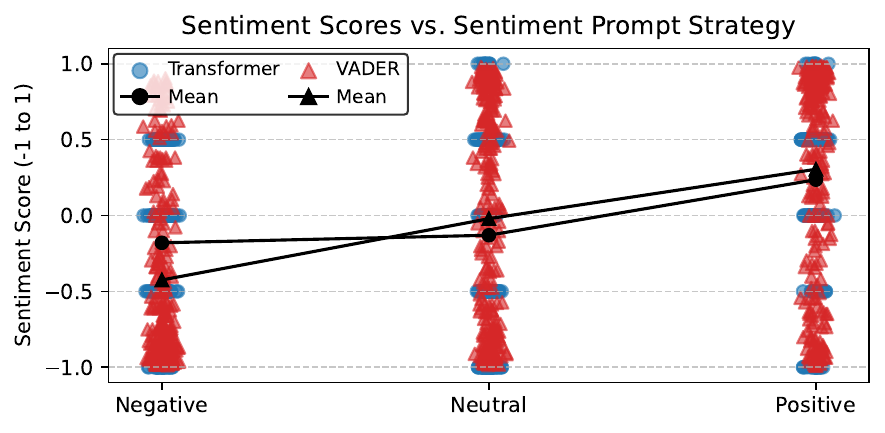}
  \includegraphics[width=0.49\linewidth]{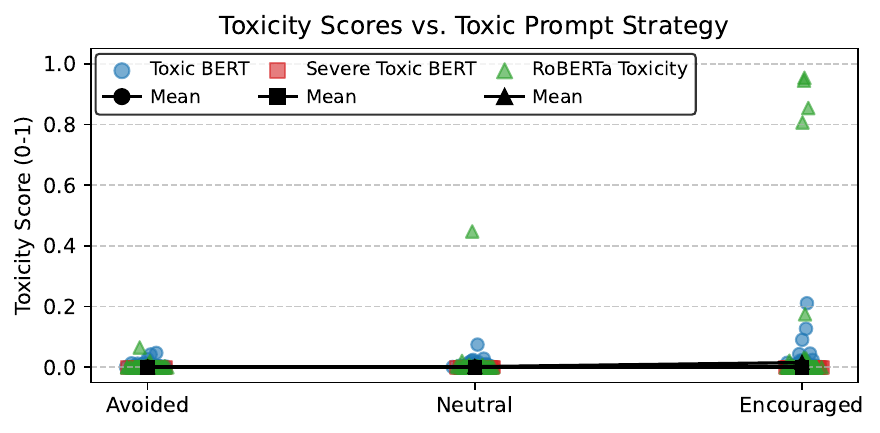} \hfill
  \includegraphics[width=0.49\linewidth]{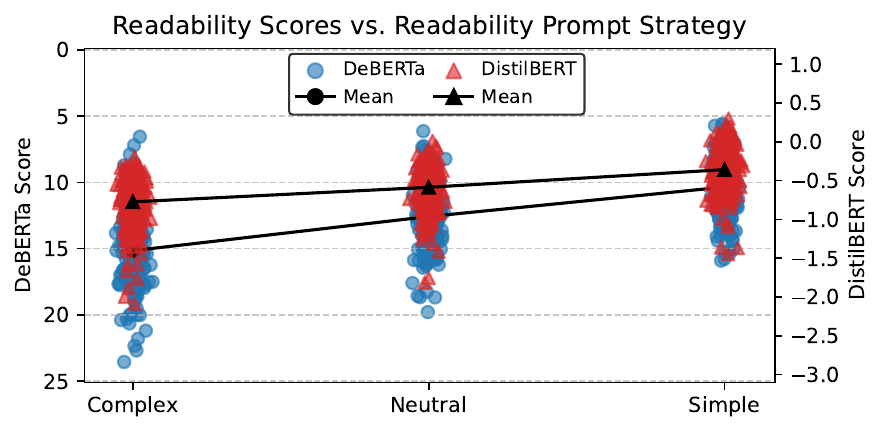}
  \vspace{-0.3cm}
  \caption {Effects of text property discouraging, neutral and encouraging prompts. Prompting for topical focus is not meaningfully effective. Prompting for sentiment has the intended effect on summary sentiment, but is not as strong as changes achieved by steering with large steering strengths. Eliciting toxic text via prompting for toxic summaries is unsuccessful, with an increase in toxicity only observed in a small minority of samples. Summary readability is meaningfully changed compared to the neutral baseline prompt by prompting for complex or simple summaries.}
  \label{fig:prompt_engineering_for_topic_and_sentiment}
\end{figure*}

\newpage
\subsection{Prompting efficacy across model scales: Llama-3.2-1B (left), Llama-3.2-3B (middle), Llama-3.1-8B (right) on NEWTS}
\label{app:prompting_effects_across_model_scales}
\begin{figure*}[!htp]
\centering
    \begin{subfigure}{\linewidth}
  \includegraphics[width=0.32\linewidth]{figures/prompt_engineering/figure_26.pdf} \hfill
  \includegraphics[width=0.32\linewidth]{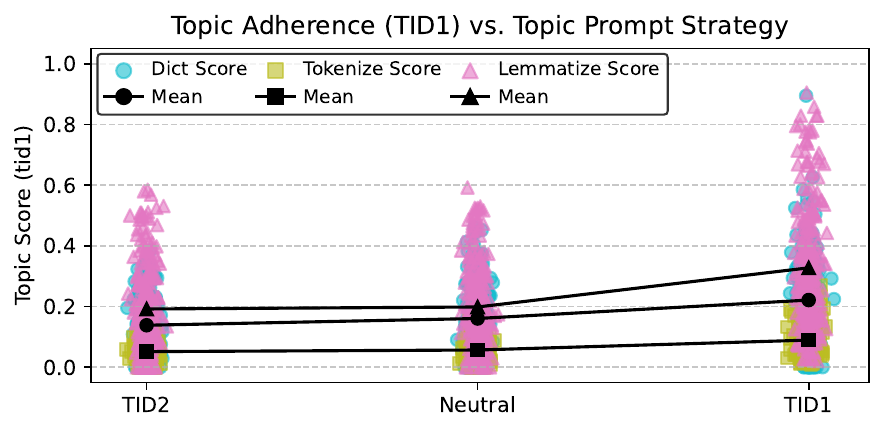} \hfill
    \includegraphics[width=0.32\linewidth]{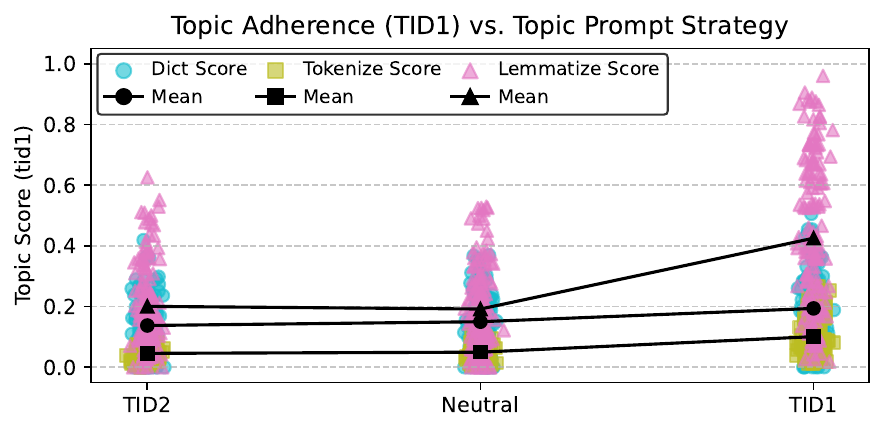}
  \subcaption{Prompting for topical focus only works for the 3B and 8B model. Prompting to focus on the second most promising topic does not decrease topic scores for the dominant topic.}
  \end{subfigure}
\vspace{0.7cm}
    \begin{subfigure}{\linewidth}
  \includegraphics[width=0.32\linewidth]{figures/prompt_engineering/figure_27.pdf} \hfill
  \includegraphics[width=0.32\linewidth]{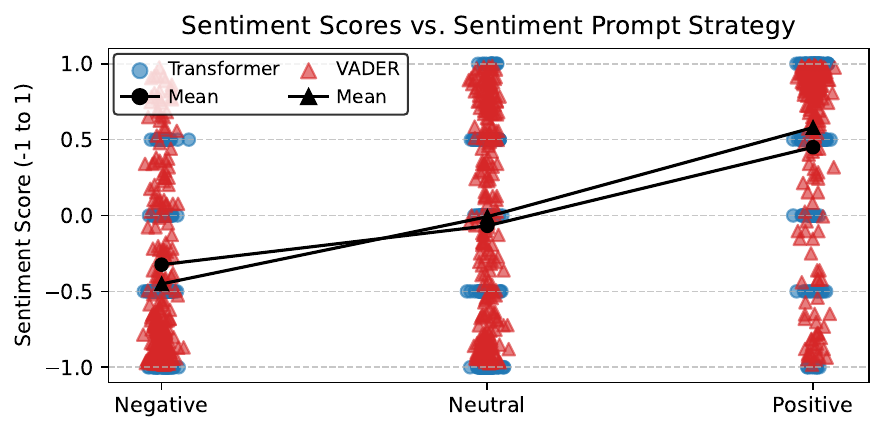} \hfill
    \includegraphics[width=0.32\linewidth]{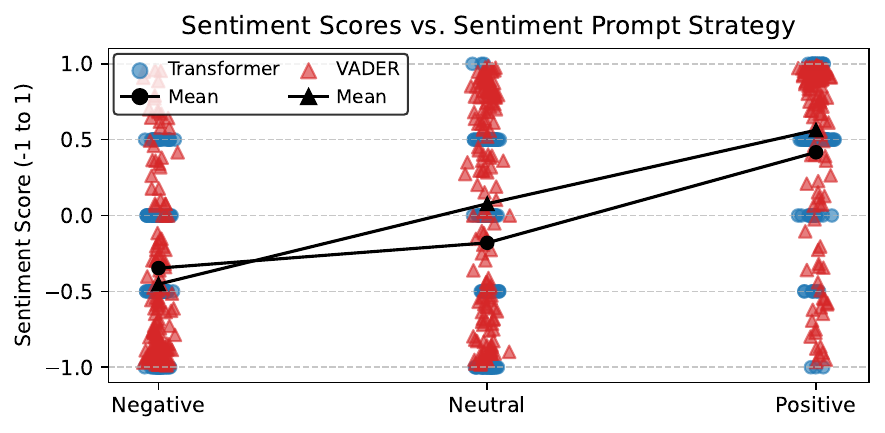}
  \subcaption{Prompting for summaries with a specific sentiment works for all model sizes. Summaries of the 3B and 8B model are more strongly influenced.}
  \end{subfigure}
\vspace{0.4cm}
    \begin{subfigure}{\linewidth}
  \includegraphics[width=0.32\linewidth]{figures/prompt_engineering/figure_28.pdf} \hfill
  \includegraphics[width=0.32\linewidth]{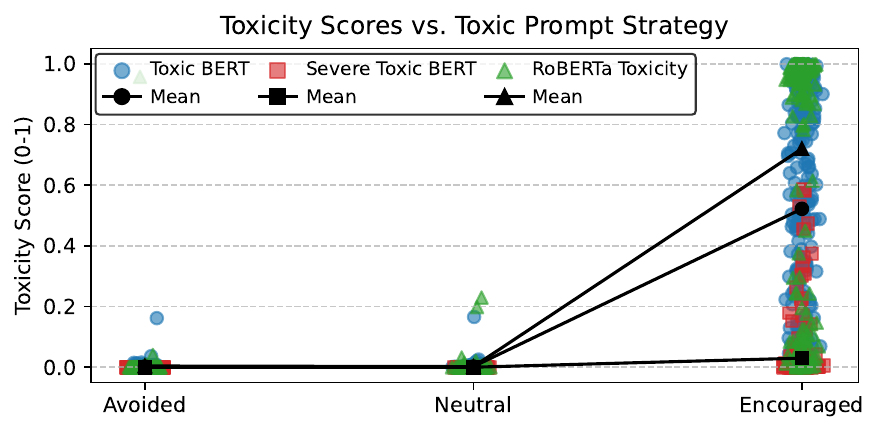} \hfill
    \includegraphics[width=0.32\linewidth]{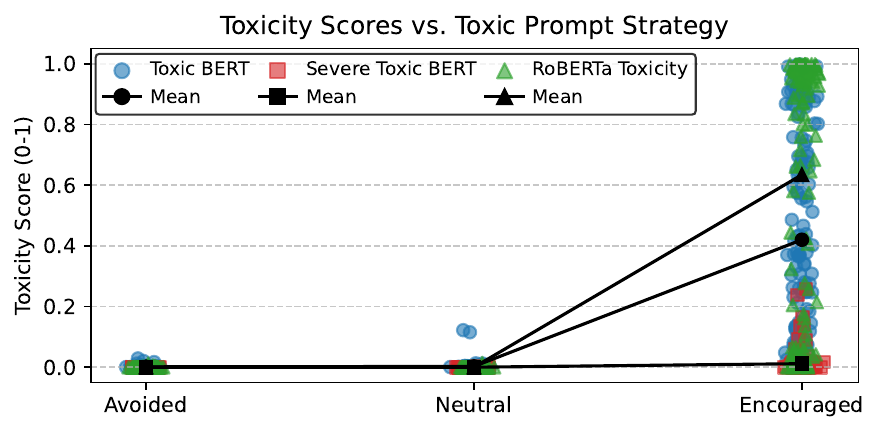}
  \subcaption{Prompting for toxic or explicitly non-toxic summaries only works for the 3B and 8B model.}
  \end{subfigure}
\vspace{0.4cm}
    \begin{subfigure}{\linewidth}
  \includegraphics[width=0.32\linewidth]{figures/prompt_engineering/figure_29.pdf} \hfill
  \includegraphics[width=0.32\linewidth]{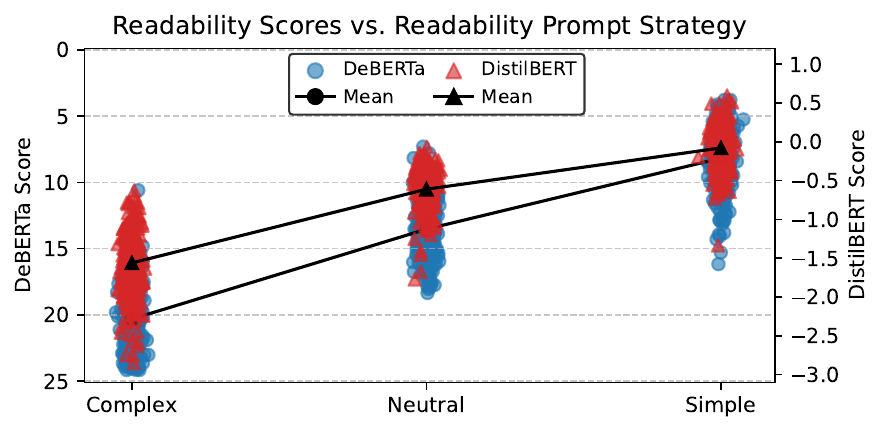} \hfill
    \includegraphics[width=0.32\linewidth]{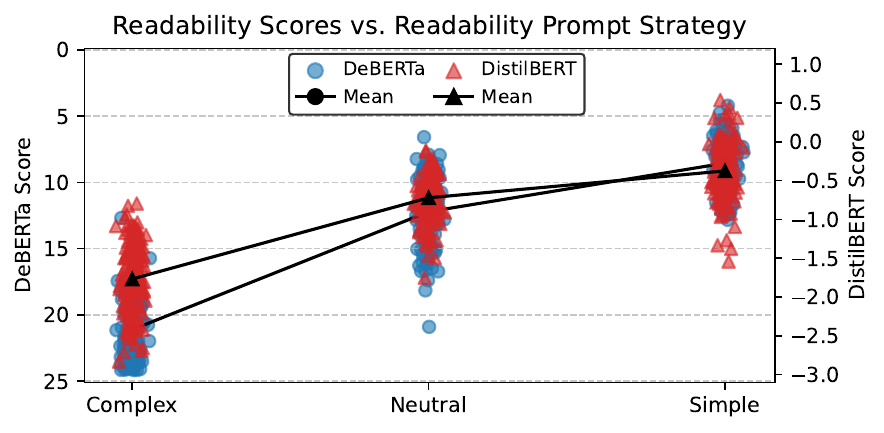}
  \subcaption{Prompting for readability has the desired impact on summaries for all model sizes, but the effect size increases with model size.}
  \end{subfigure}
\caption{Efficacy of prompting increases with model size. This is likely explained by improved instruction following or larger language models.}
\end{figure*}

\newpage

\subsection{Prompting only has minimal Effects on Text Quality}
\label{app:prompting_side_effects_on_quality}
\begin{figure*}[!htp]
\centering
\begin{subfigure}{\linewidth}
  \includegraphics[width=0.49\linewidth]{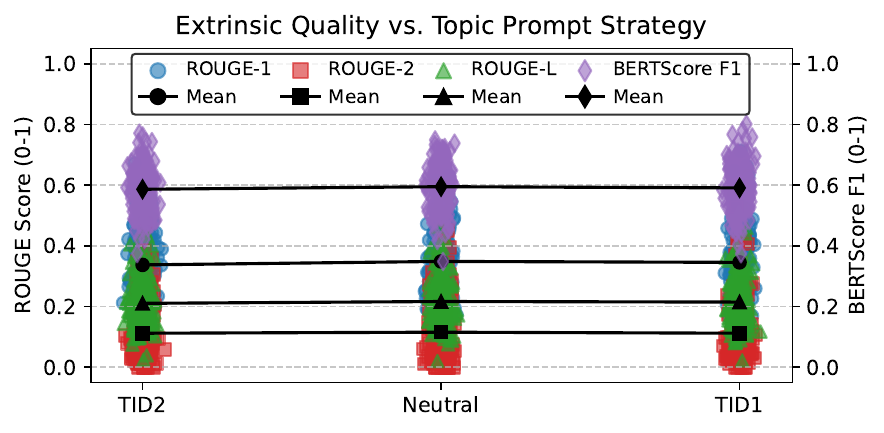} \hfill
  \includegraphics[width=0.49\linewidth]{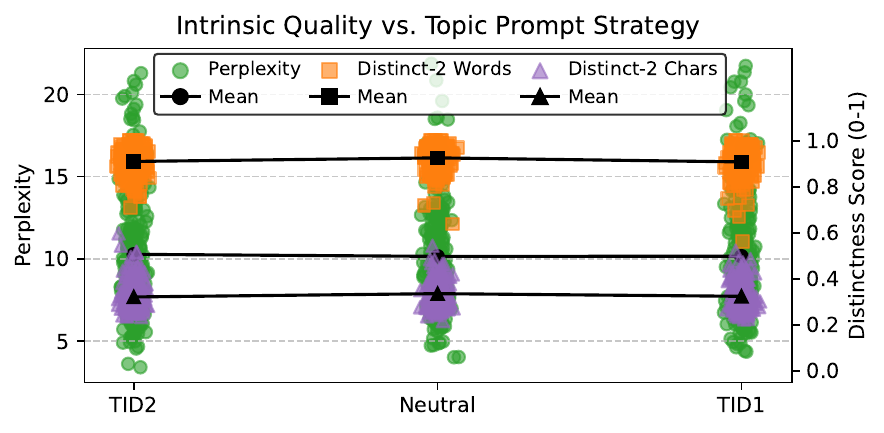}
  \subcaption {Prompting for topical focus does not meaningfully change the extrinsic quality compared to reference summaries or the intrinsic quality of the generated summaries.}
  \label{fig:topic_prompting_effect_on_text_quality}
  \end{subfigure}
\vspace{0.7cm}
\begin{subfigure}{\linewidth}
  \includegraphics[width=0.49\linewidth]{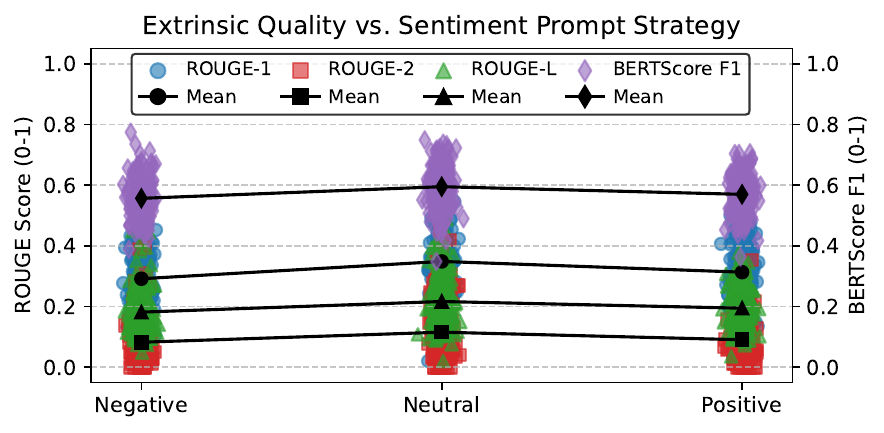} \hfill
  \includegraphics[width=0.49\linewidth]{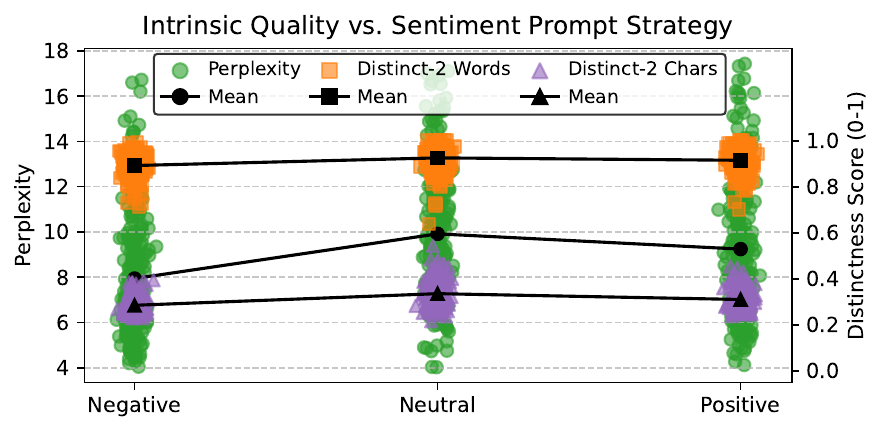}
  \subcaption {Steering for sentiment marginally reduces the extrinsic quality. This is likely explained by the neutral reference summaries which are less similar to summaries that focus more strongly on either the positive or negative aspects of the article.}
  \label{fig:sentiment_prompting_effect_on_text_quality}
  \end{subfigure}
\vspace{0.4cm}
    \begin{subfigure}{\linewidth}
  \includegraphics[width=0.49\linewidth]{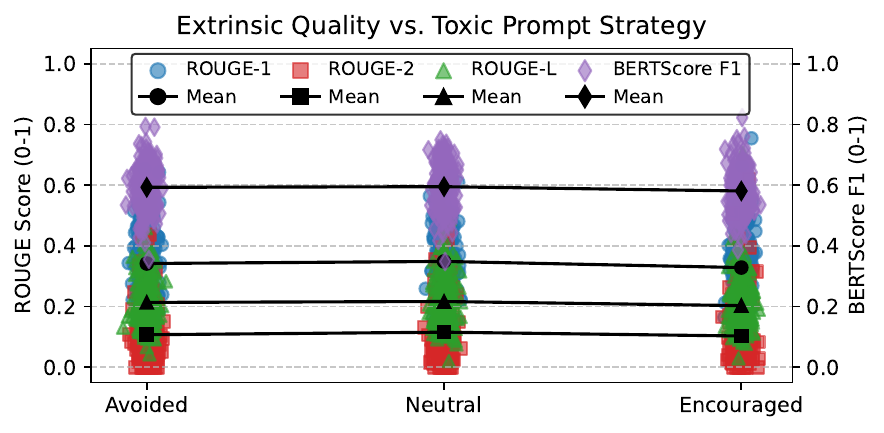} \hfill
  \includegraphics[width=0.49\linewidth]{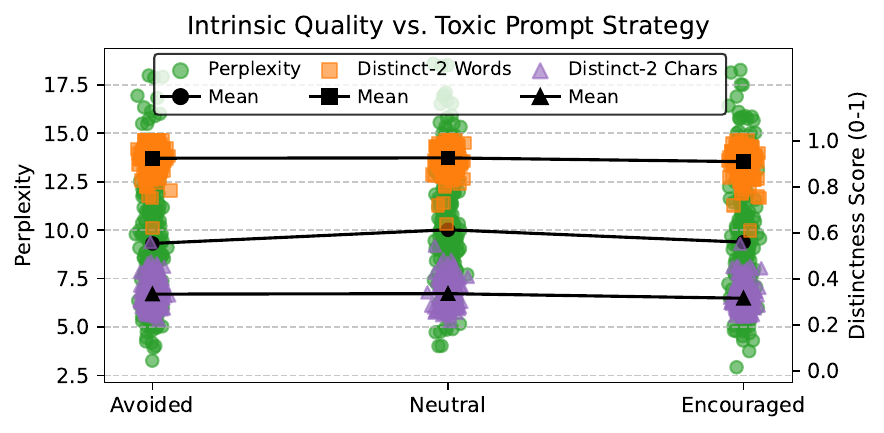}
  \subcaption {Prompting for toxic or explicitly non-toxic summaries does not meaningfully impact extrinsic or intrinsic quality. Prompting for toxicity also does not meaningfully impact the toxicity of generated summaries.}
  \label{fig:toxicity_prompting_effect_on_text_quality}
  \end{subfigure}
\vspace{0.4cm}
\begin{subfigure}{\linewidth}
  \includegraphics[width=0.49\linewidth]{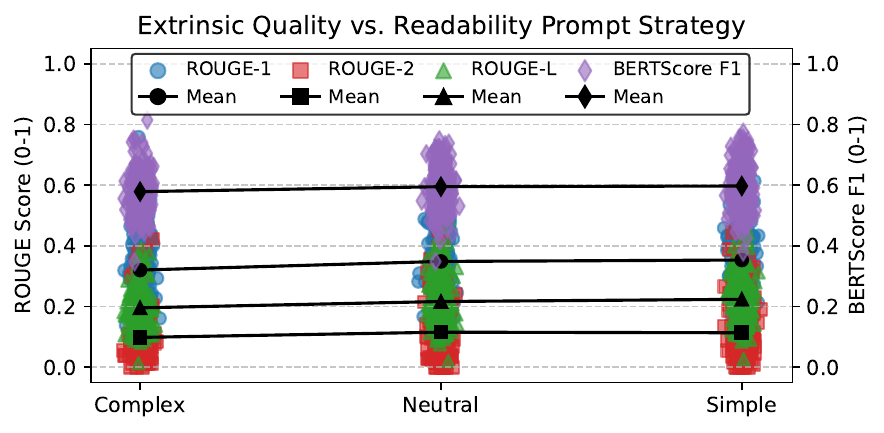} \hfill
  \includegraphics[width=0.49\linewidth]{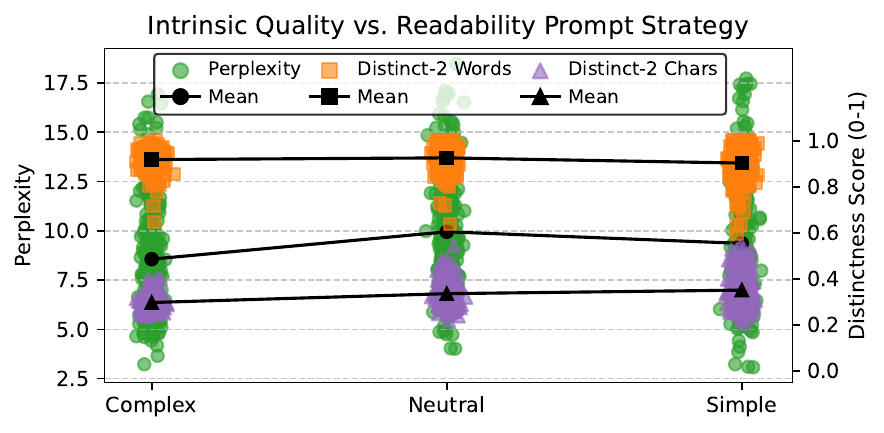}
  \subcaption {Prompting for easier readability marginally improves the measured extrinsic quality and similarity to the reference summaries. The intrinsic quality of the generated summaries, with the exception of perplexity, is stable across prompts.}
  \label{fig:readability_prompting_effect_on_text_quality}
  \end{subfigure}
\end{figure*}

\newpage
\subsection{Prompting does not meaningfully impact unrelated properties}
\begin{figure*}[!htp]
\centering
\begin{subfigure}{\linewidth}
  \includegraphics[width=0.49\linewidth]{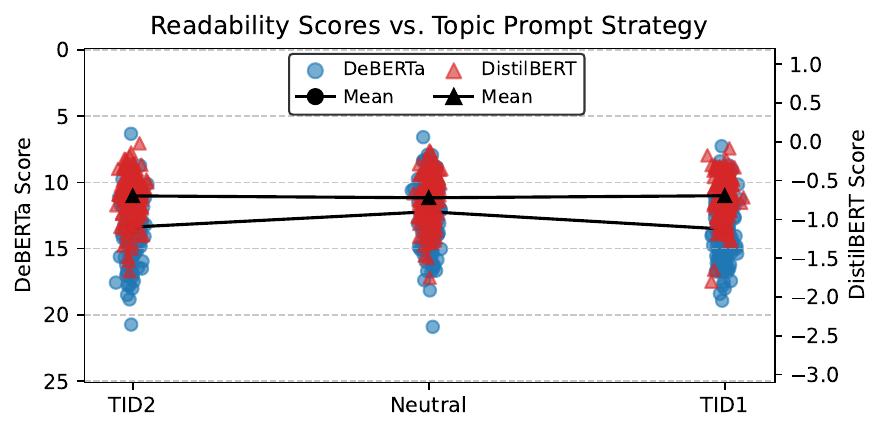} \hfill
  \includegraphics[width=0.49\linewidth]{figures/steering_impact_on_unrelated_properties/figure_18.pdf}
  \vspace{-0.1cm}
  \subcaption {Topic prompting does not meaningfully change readability or sentiment scores.}
  \label{fig:topic_prompting_impact_on_unrelated_properties}
  \end{subfigure}
\vspace{0.1cm}
\begin{subfigure}{\linewidth}
  \includegraphics[width=0.49\linewidth]{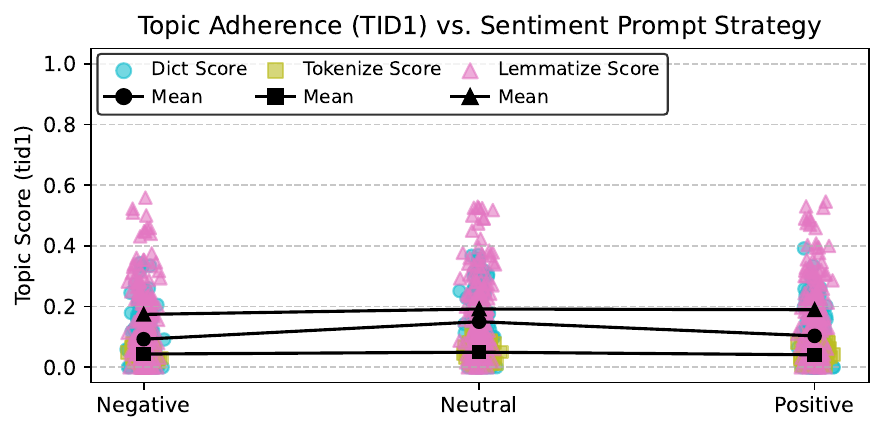} \hfill
  \includegraphics[width=0.49\linewidth]{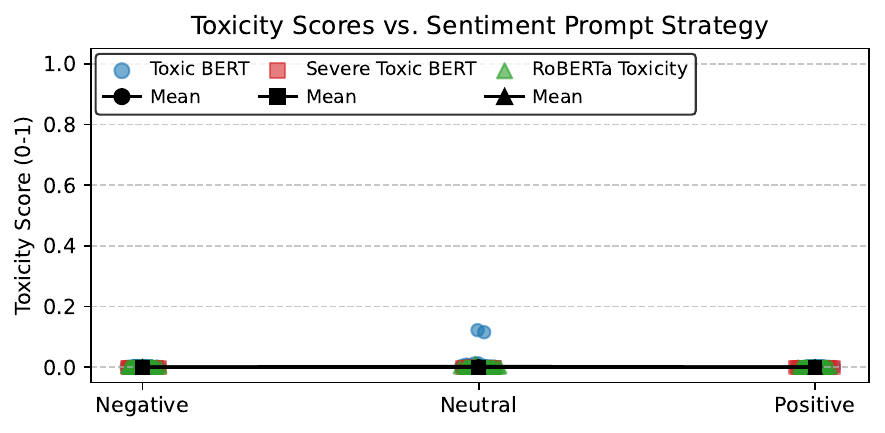}
\vspace{-0.1cm}
  \subcaption {Sentiment prompting does not meaningfully change topic or toxicity scores.}
  \label{fig:sentiment_prompting_impact_on_unrelated_properties}
  \end{subfigure}
\vspace{0.1cm}
\begin{subfigure}{\linewidth}
  \includegraphics[width=0.49\linewidth]{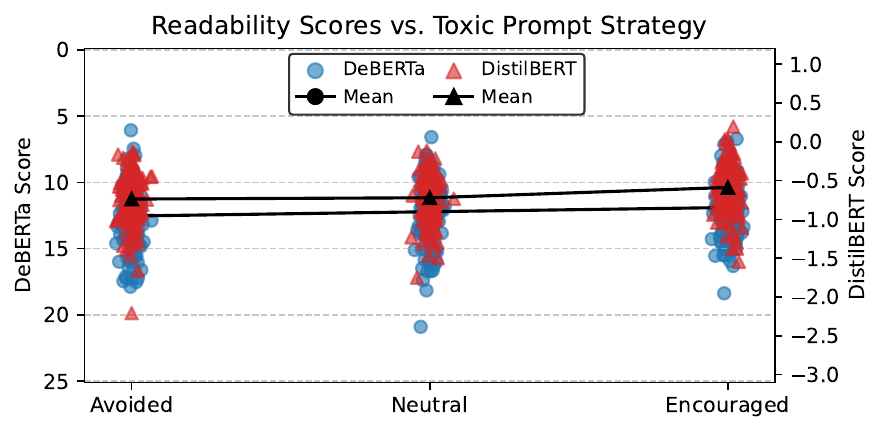} \hfill
  \includegraphics[width=0.49\linewidth]{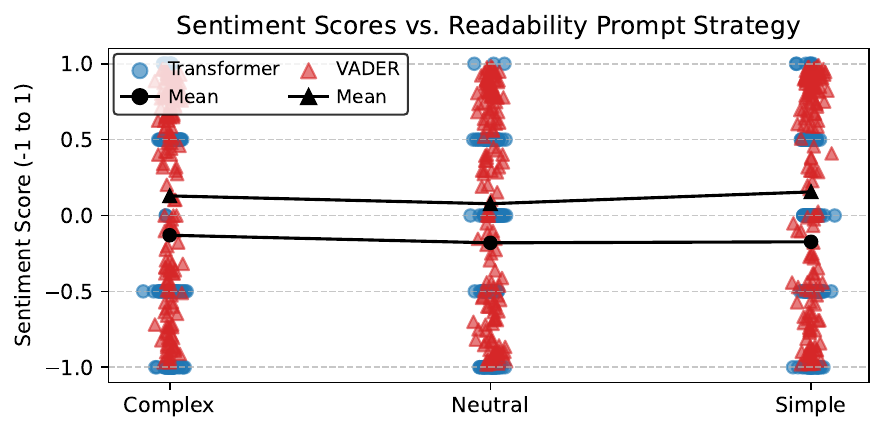}
  \subcaption {Toxicity prompting does not meaningfully change readability or sentiment scores.}
  \vspace{-0.1cm}
  \label{fig:toxicity_prompting_impact_on_unrelated_properties}
  \end{subfigure}
\vspace{0.1cm}
\begin{subfigure}{\linewidth}
  \includegraphics[width=0.49\linewidth]{figures/prompting_impact_on_unrelated_properties/figure_50.pdf} \hfill
  \includegraphics[width=0.49\linewidth]{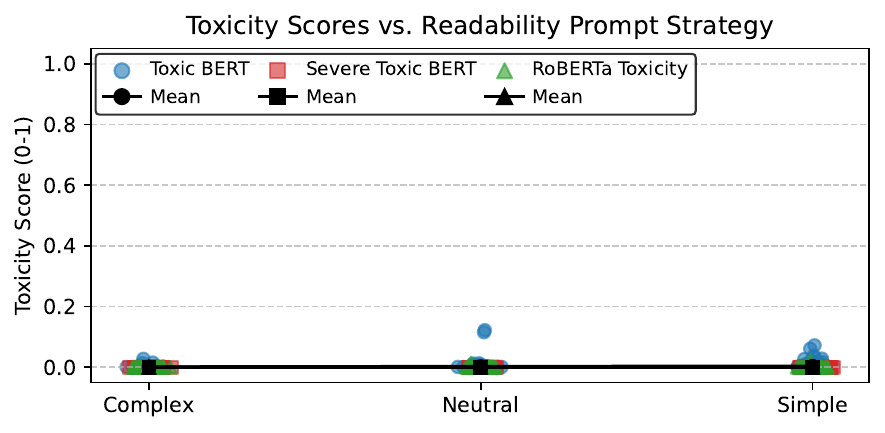}
  \subcaption {Readability prompting does not meaningfully change sentiment or toxicity scores.}
  \label{fig:readability_prompting_impact_on_unrelated_properties}
  \end{subfigure}
\vspace{-0.1cm}
\caption{Results are shown of Llama-3.1-8B, but are similar for the smaller 1B, 3B and 70B Llama models as well as the Qwen and Gemma models. Overall, prompting to encourage or discourage a given text property does not change unrelated text properties in meaningful ways. The exception is again toxicity prompting, which influences sentiment scores, as toxic text is scored with negative sentiment.}
\label{fig:prompting_impact_on_unrelated_properties}
\end{figure*}

\newpage
\subsection{Comparing Steering to Combined Steering and Prompt Engineering}
\label{app:comparing_steering_to_combined_steering_and_prompt_engineering}
\begin{figure*}[!htp] 
  \centering
  \begin{subfigure}{\linewidth}
    \centering
    \includegraphics[width=0.49\linewidth]{figures/topic_steering/figure_03.pdf} \hfill
    \includegraphics[width=0.49\linewidth]{figures/steering_and_prompting/figure_10.pdf}
    \subcaption{Combined topical prompting and steering outperforms steering across all steering strengths. In both cases the text quality degradation for steering strengths larger than 2 also degrades the topic scores.}
    \label{fig:comparing_topical_control_combined_vs_only_steering}
  \end{subfigure}
  \vspace{0.2cm} 
  \begin{subfigure}{\linewidth}
    \centering
    \includegraphics[width=0.49\linewidth]{figures/sentiment_steering/figure_01.pdf} \hfill
    \includegraphics[width=0.49\linewidth]{figures/steering_and_prompting/figure_13.pdf}
    \subcaption{Combined sentiment steering and prompting outperforms steering, especially for lower steering magnitudes. Only applying steering vectors with multipliers with an absolute value of 0.5 only shifts the sentiment by less than 0.25. If combined with prompting the change for the same steering strength more than doubles.}
    \label{fig:comparing_sentiment_control_combined_vs_only_steering}
  \end{subfigure}
  \vspace{0.2cm} 
  \begin{subfigure}{\linewidth}
    \centering
    \includegraphics[width=0.49\linewidth]{figures/toxicity_steering/figure_02.pdf} \hfill
    \includegraphics[width=0.49\linewidth]{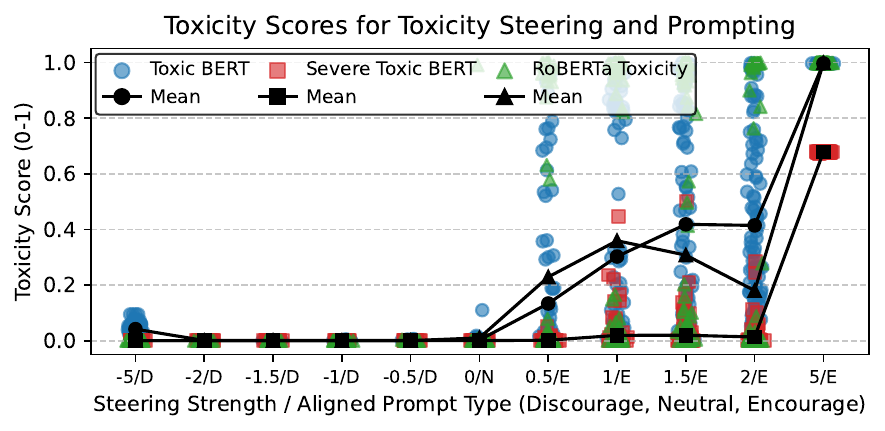}
    \subcaption{Amplifying toxicity steering with toxicity encouraging prompting greatly increases toxic output for any $\lambda > 0$. Toxicity steering alone requires $\lambda > 1.5$ to achieve a meaningful proportion of toxic summaries.}
    \label{fig:comparing_toxicity_control_combined_vs_only_steering}
  \end{subfigure}
  \vspace{0.2cm} 
  \begin{subfigure}{\linewidth}
    \centering
    \includegraphics[width=0.49\linewidth]{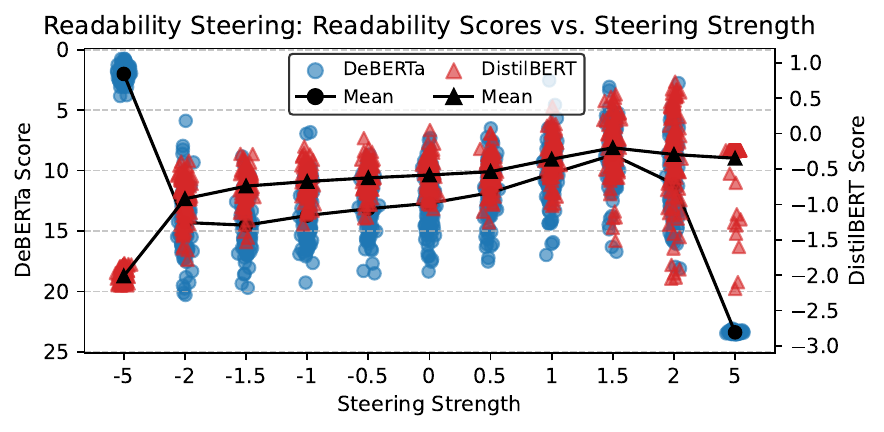} \hfill
    \includegraphics[width=0.49\linewidth]{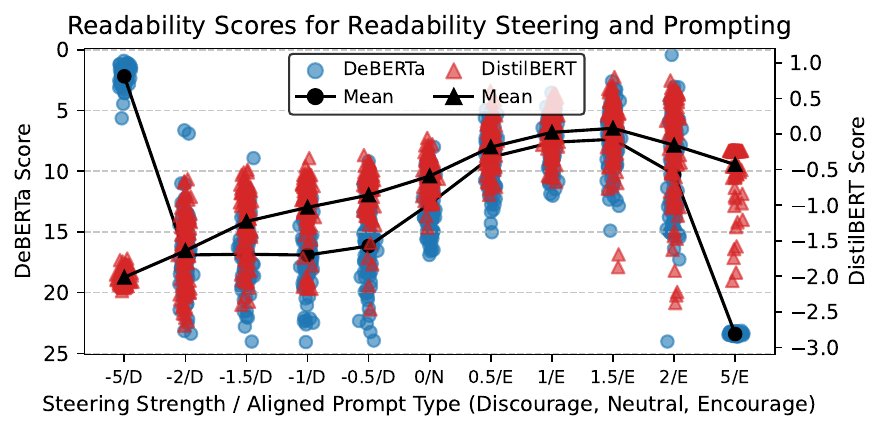}
    \subcaption{Combining readability prompting with readability steering visibly increases the effect size both by making summaries simpler or more complex, depending on the methods target direction.}
    \label{fig:comparing_readability_control_combined_vs_only_steering}
  \end{subfigure}
  \caption{Overall comparison of steering vs. combined steering and prompt engineering across different aspects.}
  \label{fig:steering_vs_steering_and_prompting_comparison}
\end{figure*}

\newpage
\subsection{Combined prompting and steering efficacy across model scales: Llama-3.2-1B (left), Llama-3.2-3B (middle), Llama-3.1-8B (right)}
\label{app:combined_steering_and_prompting_effects_across_model_scales}
\vspace{-0.4cm}
\begin{figure*}[!htp]
\centering
    \begin{subfigure}{\linewidth}
  \includegraphics[width=0.49\linewidth]{figures/steering_and_prompting/figure_10.pdf} \hfill
  \includegraphics[width=0.49\linewidth]{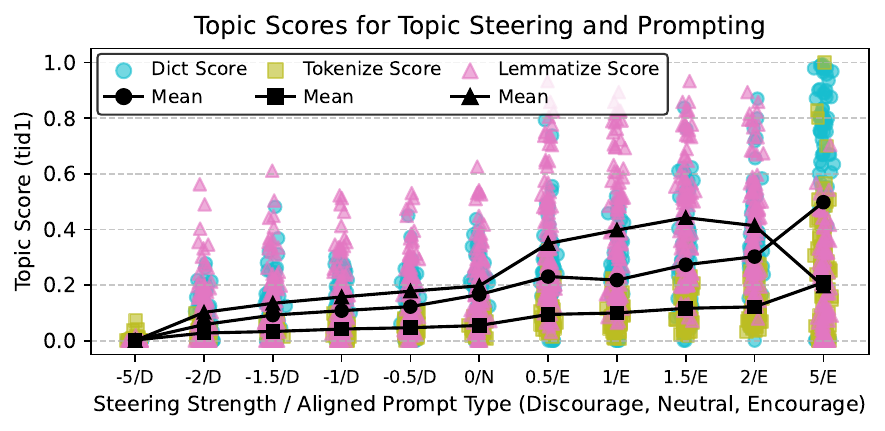} 
  \subcaption{The changes in topical focus follow a similar pattern across model sizes. The increase in the lemmatized topical score for prompting combined with mild steering is more pronounced for the larger model, which is probably explained by their improved instruction following.}
  \end{subfigure}
\vspace{0.7cm}
    \begin{subfigure}{\linewidth}
  \includegraphics[width=0.49\linewidth]{figures/steering_and_prompting/figure_13.pdf} \hfill
  \includegraphics[width=0.49\linewidth]{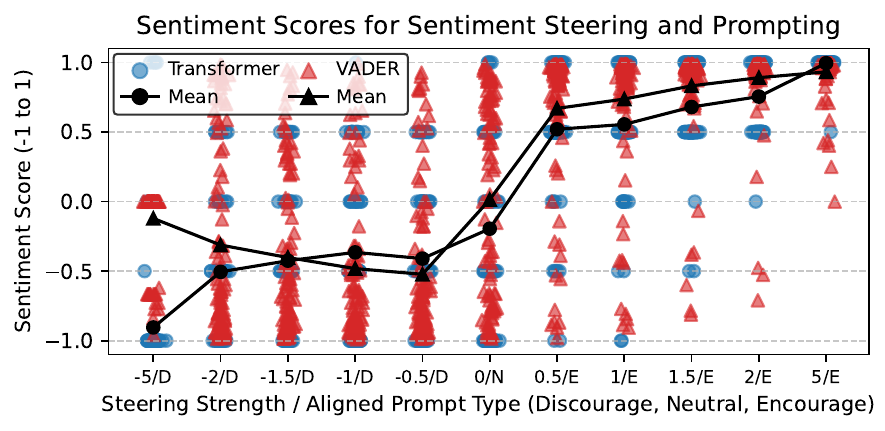} 
  \subcaption{The resulting sentiment scores of the generated summaries follow the same pattern. Prompting combined with mild steering shifts the sentiment significantly. Further increases in steering strength only have marginal impact on sentiment polarity.}
  \end{subfigure}
\vspace{0.1cm}
    \begin{subfigure}{\linewidth}
  \includegraphics[width=0.49\linewidth]{figures/steering_and_prompting/figure_52.pdf} \hfill
  \includegraphics[width=0.49\linewidth]{figures/steering_and_prompting/figure_11.pdf} 
  \subcaption{The efficacy on influencing toxicity improves with increased model size.}
  \end{subfigure}
\vspace{0.4cm}
    \begin{subfigure}{\linewidth}
  \includegraphics[width=0.49\linewidth]{figures/steering_and_prompting/figure_54.pdf} \hfill
  \includegraphics[width=0.49\linewidth]{figures/steering_and_prompting/figure_12.pdf} 
  \subcaption{Combined steering and prompting have a larger effect on readability, both for increasing or decreasing readability. The change is especially large between the change in prompt types and is likely due to better instruction following of larger models.}
  \end{subfigure}
  \vspace{-0.6cm}
\caption{Increased language model scale improves efficacy of combined steering and prompting.}
  \label{fig:comined_steering_and_prompting_effect_across_model_scale}
\end{figure*}
\newpage
\subsection{Side Effects of Combining Steering Vectors and Prompt Engineering}
\label{app:side_effects_of_combining_steering_and_prompting}
\begin{figure*}[!htp] 
  \centering
  \begin{subfigure}{\linewidth}
    \centering
    \includegraphics[width=0.49\linewidth]{figures/steering_and_prompting/figure_14.pdf} \hfill
    \includegraphics[width=0.49\linewidth]{figures/steering_and_prompting/figure_15.pdf}
    \subcaption{Combined steering and prompting for topical focus negatively impacts extrinsic and intrinsic quality for steering magnitudes $|\lambda| > 1$. Nevertheless, it enables stronger topical focus than steering or prompting alone with minimal degradation at lower $\lambda$ values.}
  \end{subfigure}
  \vspace{0.2cm} 
  \begin{subfigure}{\linewidth}
    \centering
    \includegraphics[width=0.49\linewidth]{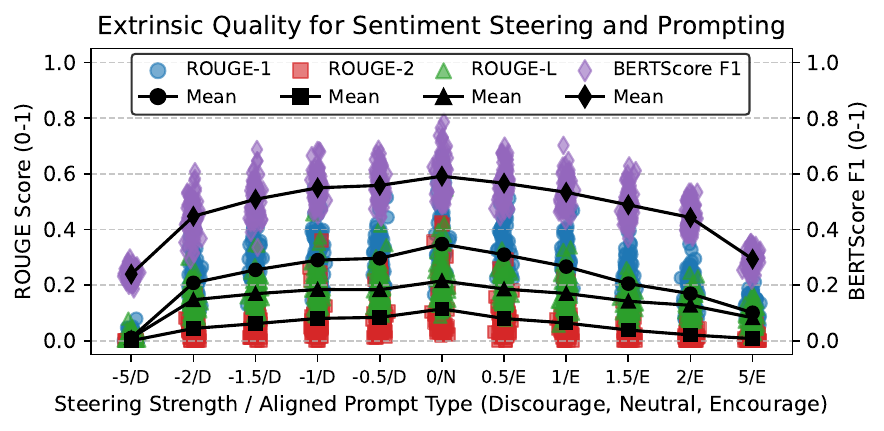} \hfill
    \includegraphics[width=0.49\linewidth]{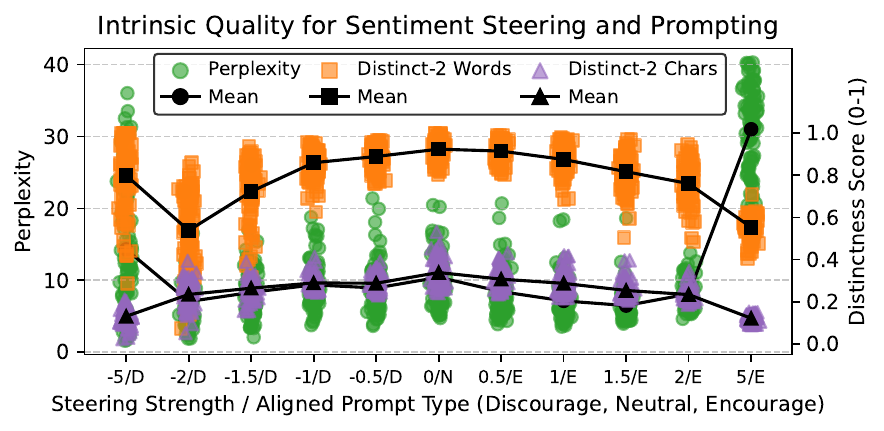}
    \subcaption{Using hybrid sentiment control incurs minor but observable text quality costs. Given that small values of the steering strength $\lambda$ produce large sentiment changes, effective control with minimal quality degradation is feasible.}
  \end{subfigure}
  \vspace{0.2cm} 
  \begin{subfigure}{\linewidth}
    \centering
    \includegraphics[width=0.49\linewidth]{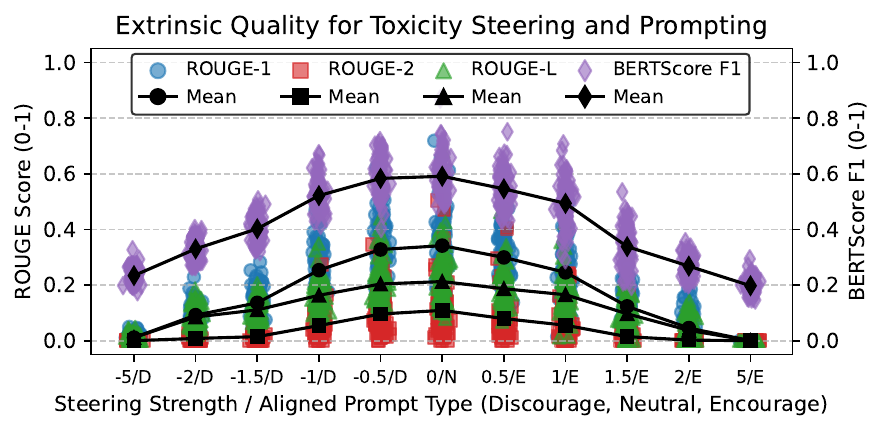} \hfill
    \includegraphics[width=0.49\linewidth]{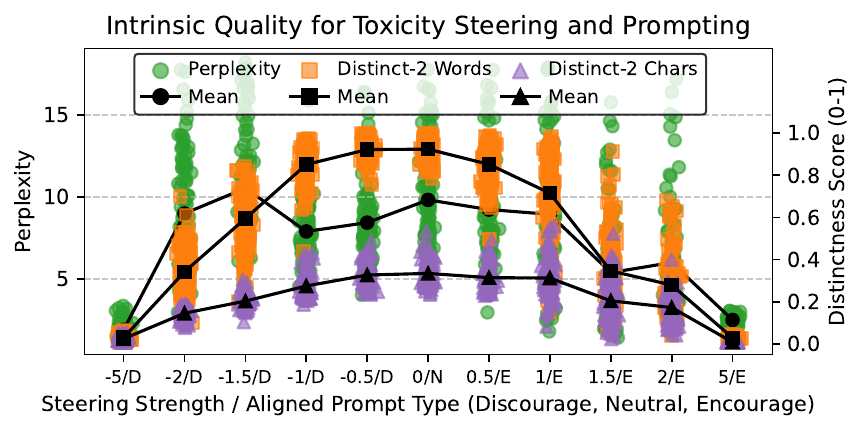}
    \subcaption{As for steering vectors alone, the hybrid approach for toxicity control most severely impacts text quality. For steering strengths $\lambda \geq 1.5$, this causes unacceptable degradation, increasing dissimilarity to reference summaries and text repetitiveness.}
  \end{subfigure}
  \vspace{0.2cm} 
  \begin{subfigure}{\linewidth}
    \centering
    \includegraphics[width=0.49\linewidth]{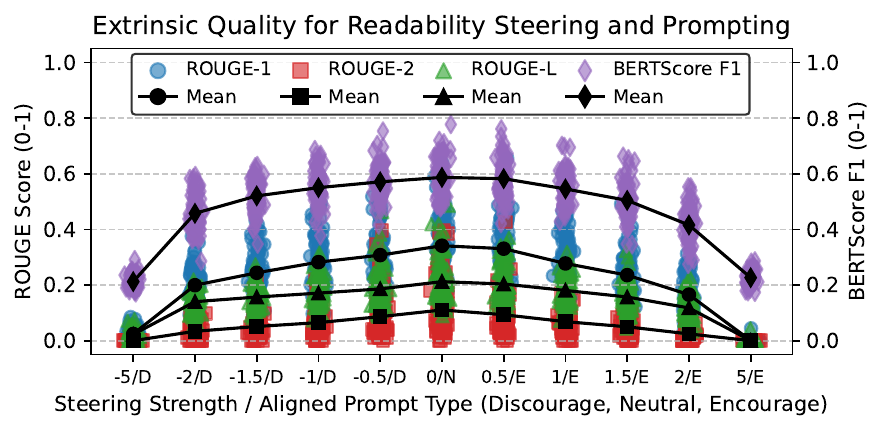} \hfill
    \includegraphics[width=0.49\linewidth]{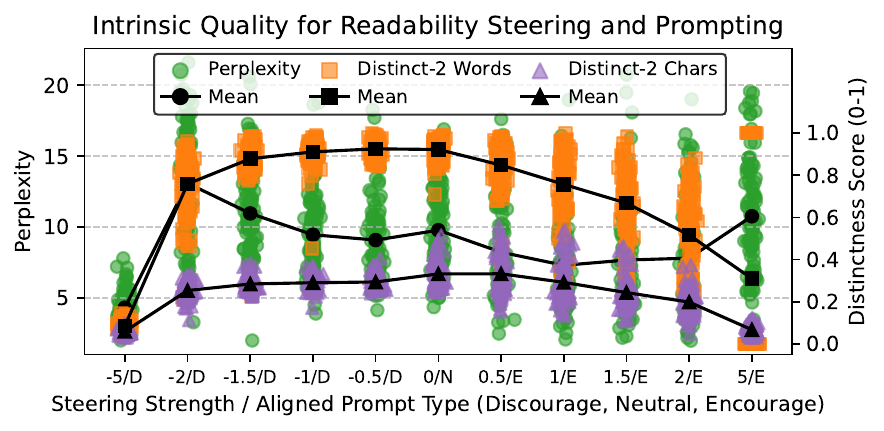}
    \subcaption{Steering and prompting for readability mildly affects extrinsic text quality for moderate steering strengths. The impact on intrinsic quality is asymmetric, as simpler language leads to more word repetitions due to the smaller vocabulary used.}
  \end{subfigure}
  \vspace{-0.5cm}
  \caption{Combined steering and prompting offers a better efficacy-quality tradeoff than steering or prompting alone. Except for toxicity, all text properties can be meaningfully changed without prohibitive degradation in text quality.}
  \label{fig:side_effects_of_steering_and_prompting}
\end{figure*}

\newpage
\subsection{Individual examples}
\label{app:individual_samples}

\subsubsection{Sentiment Steering Summary Example}
\begin{table*}[htbp]
\centering
\caption{Summaries were generated with the Qwen 3 32b model for the same article on the SAMSum dataset, but steered with different strengths towards negative sentiment ($\lambda = -1$), no steering ($\lambda = 0$) and positive sentiment ($\lambda = 1$). Negative sentiment is color-coded in \textcolor{red}{red}, positive sentiment in \textcolor{green}{green}.}
\label{tab:sentiment_examples_qwen32b}
\begin{tabular}{|p{0.3\textwidth}|p{0.3\textwidth}|p{0.3\textwidth}|}
\hline
\textbf{Steering strength $\lambda = -1$} & \textbf{Steering strength $\lambda = 0$} & \textbf{Steering Strength $\lambda = 1$} \\
\hline
John grapples with escalating stress from homework (pages 15, exercises 2-3, plus repetitive chapter reading) and a \textcolor{red}{stifling work environment, where his boss’s harsh demands and colleagues’ passive compliance fuel frustration}. He doubts his ability to manage the workload while feeling isolated and \textcolor{red}{resentful}. Though Cassandra offers reassurance and casual \textcolor{green}{support}, the plan for drinks \textcolor{red}{barely offsets his sense of helplessness and simmering resentment}. & John asks Cassandra about homework for the next day, and she confirms it includes page 15 exercises 2 and 3, additional chapter reading, and vocabulary review. He vents about his \textcolor{red}{toxic} work environment, and Cassandra offers support while they plan to meet for drinks later in the week. Cassandra \textcolor{green}{reassures} him about the homework and emphasizes her availability as a teacher. & John seeks help from Cassandra regarding homework (pages 15, exercises 2-3, plus chapter reading and vocabulary review) and shares work \textcolor{red}{frustrations}, but she calmly offers support and clarification. Despite challenges at work, he remains engaged and open to solutions, while Cassandra \textcolor{green}{reinforces his confidence by affirming her role as a teacher and willingness to assist}. Their plan for drinks later in the week adds a \textcolor{green}{friendly, hopeful touch, highlighting teamwork and mutual encouragement}.
\\
\hline
\end{tabular}
\end{table*}

\newpage
\begin{table*}[htbp]
\centering
\caption{Summaries were generated with the Llama-3.2-1B model for the same NEWTS article, but steered with different strengths towards negative sentiment ($\lambda = -1.5$), no steering ($\lambda = 0$) and positive sentiment ($\lambda = 1.5$). Negative sentiment is color-coded in \textcolor{red}{red}, positive sentiment in \textcolor{green}{green}.}
\label{tab:sentiment_examples_llama1b}
\begin{tabular}{|p{0.3\textwidth}|p{0.3\textwidth}|p{0.3\textwidth}|}
\hline
\textbf{Steering strength $\lambda = -1.5$} & \textbf{Steering strength $\lambda = 0$} & \textbf{Steering Strength $\lambda = 1.5$} \\
\hline
The airline Finnair has been \textcolor{red}{accused} of flying in Ukraine airspace, causing \textcolor{red}{widespread outrage} and \textcolor{red}{damage to its reputation}. The airline claims that it has no flights in Ukraine and has advised that it will continue to fly in Western Ukrainian airspace, where the \textcolor{red}{crash} of the \textcolor{red}{MH17 disaster} occurred. Finnair has \textcolor{red}{apologized for the lack of flights} in Ukraine, and has \textcolor{green}{promised to put the safety of its passengers above its desire} to avoid flying over Ukraine. The airline has also suggested that the \textcolor{red}{disaster may be caused} by the lack of air travel in the region and the \textcolor{red}{failure of the pilots} to fly by the safety of the pilots, and that the \textcolor{red}{crash may be caused} by a \textcolor{red}{``safety crisis''} in the region. The situation is seen as a \textcolor{red}{major crisis} for the ...
&
Finnair has been \textcolor{red}{forced to backtrack} on its claims that it didn't fly in Ukrainian airspace after being \textcolor{red}{criticized} on social media for its handling of the Malaysian Airlines \textcolor{red}{tragedy}. The airline has admitted that it did fly over Ukraine and \textcolor{red}{apologized for any inconvenience} it may have caused to its passengers. Finnair's decision not to fly in Ukrainian airspace has been \textcolor{red}{met with criticism}, with many \textcolor{red}{accusing} the airline of being \textcolor{red}{insensitive and inconsiderate}. The airline has explained that it has made the decision to avoid flying in Ukrainian airspace due to Eurocontrol's recommendations and the need to \textcolor{green}{ensure the safety of its passengers}. The airline has \textcolor{red}{apologized for any inconvenience} it may have caused and \textcolor{green}{promises to evaluate its situation carefully} and consider the recommendations of the aviation authorities when making any further decisions.
&
Finnair, a Finnish airline, was \textcolor{green}{praised} for its \textcolor{green}{efforts to reassure passengers} about the \textcolor{green}{safety} of their flights after the \textcolor{red}{MH17 disaster}. They were able to share information about the airline's flights in Ukraine and even took to social media to \textcolor{green}{express their sympathy} to their fellow oneworld partner, Singapore Airlines. However, they were also quick to admit that they didn't fly over Ukraine airspace, which was a \textcolor{green}{great relief for passengers}. The airline also \textcolor{green}{extended an apology} for any \textcolor{red}{inconvenience} caused by their flight in Ukraine. The article also notes that the airline has continued to fly in Ukraine airspace and has made the decision to not fly in Ukraine airspace for time being, which is a \textcolor{green}{great relief for passengers}. The airline explained that their flights in Ukraine airspace will be ...
\\
\hline
\end{tabular}
\end{table*}

\newpage
\subsubsection{Combined Steering and Prompting for Readability}
\begin{table*}[!ht]
\centering
\caption{Examples of readability steering combined with readability prompting applied to the Gemma 3 27b model, demonstrating varied outputs for the same source text from the arXiv dataset under different steering conditions. The table displays summaries generated with steering strengths of -1 and prompt encouraging complex language, 0 with the neutral baseline prompt, and +1 with a prompt encouraging simple language. Within these generated texts, \textcolor{red}{complex words and phrases are highlighted in red}, while \textcolor{green}{simpler phrasings are highlighted in green} to visually illustrate the impact of steering on language characteristics.}
\label{tab:readability_examples_gemma27b}
\begin{tabular}{|p{0.3\textwidth}|p{0.3\textwidth}|p{0.3\textwidth}|}
\hline
\textbf{$\lambda = -1$ + complexity prompting} & \textbf{$\lambda = 0$ + neutral prompting} & \textbf{$\lambda = +1$ + simplicity prompting} \\
\hline
This investigation centers on the \textcolor{red}{asymptotic behavior of regularized kernel-based algorithms} operating within the representational framework of additive models, seeking to \textcolor{red}{delineate conditions} under which an additive kernel architecture \textcolor{red}{demonstrably surpasses} the performance characteristics of conventional, non-partitioned kernels in high-dimensional feature spaces. Utilizing a \textcolor{red}{convex and Lipschitz continuous loss function}, the research \textcolor{red}{eschews the limitations inherent in differentiable}, albeit statistically fragile, least-squares formulations. The authors establish novel learning rate bounds for these algorithms, with particular emphasis on \textcolor{red}{quantile regression employing the pinball loss}, revealing opportunities for enhanced convergence properties. A methodological contribution lies in the introduction of an intermediate functional approximation, \textcolor{red}{mitigating the challenges} associated with estimating approximation error within the additive context...
&
This paper investigates learning rates for Support Vector Machines (SVMs) within the framework of additive models, aiming to determine if an \textcolor{red}{additive kernel can outperform a general kernel} (like a Gaussian RBF) in high-dimensional settings when the additive model assumption holds. The research focuses on \textcolor{red}{regularized kernel-based methods} using a \textcolor{red}{convex and Lipschitz continuous loss function}, \textcolor{green}{moving beyond traditional} least squares loss which lacks robustness. The authors establish learning rates for these algorithms, with a particular focus on quantile regression using the pinball loss function, showing potential for improved convergence. A key novelty is addressing the difficulty of estimating approximation error in additive models by introducing an intermediate function for better component function approximation. The proposed approach can \textcolor{green}{achieve faster learning rates} than existing SVM methods in high dimensions...
&
This research looks at a way to \textcolor{green}{improve machine learning} when data can be \textcolor{green}{broken down into simpler parts}. It explores using “additive models” with Support Vector Machines (SVMs). These models can be \textcolor{green}{more accurate} and \textcolor{green}{easier to understand} than other complex methods. The study focuses on making these models \textcolor{green}{work well even with noisy data}. It shows that when data does follow an additive pattern, using this approach can lead to \textcolor{green}{faster and more accurate learning}, especially when dealing with lots of data. A key idea is a new method for estimating how well the model is working. The results suggest this approach \textcolor{green}{beats traditional methods} in high-dimensional data, but only if the data truly has an additive structure. Essentially, if you can break down the problem into smaller pieces, this method can \textcolor{green}{learn more efficiently}.
\\
\hline
\end{tabular}
\end{table*}

\begin{table*}[!ht]
\centering
\caption{Examples of readability steering combined with readability prompting applied to the Llama-3.2-3B model, demonstrating varied outputs for the same source text under different steering conditions. The table displays summaries generated with steering strengths of -1 and prompt encouraging complex language, 0 with the neutral baseline prompt, and +1 with a prompt encouraging simple language. Within these generated texts, \textcolor{red}{complex words and phrases are highlighted in red}, while \textcolor{green}{simpler phrasings are highlighted in green} to visually illustrate the impact of steering on language characteristics.}
\label{tab:readability_examples_llama3b}
\begin{tabular}{|p{0.3\textwidth}|p{0.3\textwidth}|p{0.3\textwidth}|}
\hline
\textbf{$\lambda = -1$ + complexity prompting} & \textbf{$\lambda = 0$ + neutral prompting} & \textbf{$\lambda = +1$ + simplicity prompting} \\
\hline
The United States is \textcolor{red}{precipitating} a \textcolor{red}{global economic calamity}, owing to its failure to devise a plan to augment its debt ceiling and \textcolor{red}{avert default}, thereby \textcolor{red}{precipitating a calamitous event} that will have far-reaching consequences for the global economy. As the World Bank President, Jim Yong Kim, has warned, the US's inaction will \textcolor{red}{precipitate a cataclysmic collapse of confidence}, leading to a \textcolor{red}{precipitous decline} in economic growth, and ultimately, a devastating impact on the developing world. The International Monetary Fund and its \textcolor{red}{sister lending agency}, the World Bank, have \textcolor{red}{endeavored to persuade} US policymakers to \textcolor{red}{expedite a resolution}, \textcolor{red}{lest the nation succumbs} to an \textcolor{red}{unprecedented fiscal crisis} that would \textcolor{red}{irreparably harm} the global economy.
&
The World Bank President, Jim Yong Kim, warned that the US is just five days away from \textcolor{red}{causing a global economic disaster} unless \textcolor{green}{a plan is put in place} to raise the nation's debt limit and avoid default. The US debt ceiling deadline is approaching and Treasury Secretary Jacob Lew has warned that the government will \textcolor{red}{exhaust its borrowing authority} on Thursday. The World Bank is concerned that a default would have a disastrous impact on the developing world, which would in turn affect developed economies. The IMF has also \textcolor{red}{expressed concerns} about the \textcolor{red}{near-term risks}, including the potential for the US Federal Reserve to \textcolor{red}{scale back its stimulus program}, but has noted that the US economy is strong enough to withstand a reduction in stimulus. Meanwhile, \textcolor{green}{talks between Democratic and Republican Senate leaders have begun}, with
&
The World Bank says the US is \textcolor{green}{close to causing a big economic problem} if the \textcolor{green}{politicians don't fix} the debt limit. The US \textcolor{green}{needs to borrow money to pay its bills}, but the \textcolor{green}{government is running out of money} and might not be able to pay its debts. The World Bank says this \textcolor{green}{could make interest rates go up}, and that \textcolor{green}{could make it harder for people to borrow money} and \textcolor{green}{for the economy to grow}. The US government needs to raise the debt limit so it \textcolor{green}{can keep paying its bills}. The US is running out of time, and the World Bank says the US is just `days away' from a \textcolor{green}{big problem}. The \textcolor{green}{politicians need to work together to fix} the debt limit before it's too late.
\\
\hline
\end{tabular}
\end{table*}

\newpage
\hspace{0.1cm}
\newpage
\subsubsection{Toxicity Steering Summary Example}
\begin{table*}[!ht]
\centering
\caption{Summaries were generated with the Llama-3.2-1B model for the same article, but steered with different strengths towards toxic outputs. For $\lambda < 0$ outputs are actively steered towards non-toxic output and for $\lambda > 0$ outputs are actively steered towards \textcolor{red}{toxic output}. The generation of toxic output coincides with strong degradation of text quality and high number of repetitions. \textbf{Content warning:} Some model outputs contain offensive language, which has been partially censored.}
\label{tab:toxicity_examples}
\begin{tabular}{|p{0.1\textwidth}|p{0.25\textwidth}|p{0.22\textwidth}|p{0.19\textwidth}|p{0.1\textwidth}|}
\hline
$\lambda=-5$ & $\lambda=-1.5$ & $\lambda=0$ & $\lambda=1.5$ & $\lambda=5$ \\
\hline
and and thoughtful and and and and and and and and and and and and and and and and and and warm and ...
&
This article highlights the positive trends in the energy and spending aspects of the lives of families in the UK. The annual and significant increases in the energy levels of the homes of the families have been the results of the positive and the positive and the increasing appreciation of the warmth and the satisfaction of the natural and the personal aspects of the daily lives of the people of the UK. The positive and the increased and the happy and the growing aspects of the relationships and the interactions of the people of the lives of the people of the positive and the and the supportive and the generous expressions of the individual and the nature of the relationships and the personal aspects of the daily lives of the people of the positive and the and the and the positive and the and the supportive and the 
&
UK households have seen their energy bills soar by 55\% in a decade, despite cutting gas and electricity usage by almost a fifth. The Office for National Statistics said the increase in spending is solely due to rises in energy prices, while average household energy usage has fallen by 17\%. The UK is the most expensive region for households to live in, with Scottish households paying the highest average spend on energy (£112 a month). This is partly due to the higher average spend in Northern Ireland, where households pay substantially more. The government has urged households to reduce their consumption to cut bills, but the rise in spending is attributed solely to energy price increases. &
The article claims that the UK's energy price is 55 per cent, and the ONS figures show that the average house in the UK is now spending £105 on gas and £152 on electricity. The article claims the ONS figures are "exagger" and the ONS is "ex" \textcolor{red}{(excrement) in the toilet,}, and the article claims the ONS is "ex" (ex) in the can. \textcolor{red}{The article is a joke}. \textcolor{red}{The article is a joke}. 2012 was a year, and the ONS was a p in the can, 2002 was a p in the head, and 2007 was a p in the head in the head. 2002 was 55, and &
\textcolor{red}{fu**ing fu**ing fu**ing fu**ing fu**ing fu**ing fu**ing fu**ing fu**ing fu**ing fu**ing fu** fu**ing fu**ing fu**ing fu**ing fu**ing fu** fu**ing fu**ing fu** fu**ing fu**ing fu**ing fu**ing fu**ing fu**ing ...}\\
\hline
\end{tabular}
\end{table*}
\newpage
\begin{table*}[!ht]
\centering
\caption{Examples of readability steering combined with readability prompting, demonstrating varied outputs for the same source text under different steering conditions. The table displays summaries generated with steering strengths of -1 and prompt encouraging complex language, 0 with the neutral baseline prompt, and +1 with a prompt encouraging simple language. Within these generated texts, \textcolor{red}{complex words and phrases are highlighted in red}, while \textcolor{green}{simpler phrasings are highlighted in green} to visually illustrate the impact of steering on language characteristics.}
\label{tab:readability_examples_prince_harry}
\begin{tabular}{|p{0.35\textwidth}|p{0.3\textwidth}|p{0.25\textwidth}|}
\hline
\textbf{$\lambda = -1$ + complexity prompting} & \textbf{$\lambda = 0$ + neutral prompting} & \textbf{$\lambda =1$ + simplicity prompting} \\
\hline
Prince Harry's \textcolor{red}{ostensibly incongruous juxtaposition} from a \textcolor{red}{sedentary tenure} as a Staff Officer at Horse Guards to his \textcolor{red}{forthcoming reintegration} into the Army Air Corps, \textcolor{red}{specifically via the acquisition of proficiency} in the Lynx battlefield-support chopper, \textcolor{red}{precipitates a concomitant revitalization} of his \textcolor{red}{erstwhile aviator's inclinations}. Having cultivated a \textcolor{red}{predilection for the aerial domain} as an Apache pilot in Afghanistan, a role \textcolor{red}{subsequently relinquished} due to his \textcolor{red}{elevated station within the Royal Household}, Harry is poised to \textcolor{red}{rekindle his affinity for flight} by undertaking new training. This will \textcolor{red}{facilitate a reintegration} into the Army Air Corps, enabling him to \textcolor{red}{assume a role consonant} with his flight experience.
&
Prince Harry is set to return to the Army Air Corps to \textcolor{green}{prepare for flying helicopters}, including the Lynx battlefield-support chopper, after being stuck in a desk job for several months. The 30-year-old prince, who served as an Apache pilot in Afghanistan, will undergo training in the new year to convert to his new role. He is expected to spend six months to fully adjust to flying the Lynx, a fast helicopter with \textcolor{red}{powerful surveillance equipment}.
&
Prince Harry is \textcolor{green}{going back to fly helicopters}. He \textcolor{green}{was doing a desk job} for a while. He used to fly Apache helicopters in Afghanistan and \textcolor{green}{liked flying}. Now, he will learn to fly a new helicopter called the Lynx. It will \textcolor{green}{take him about six months to learn}. He is \textcolor{green}{excited to fly again} and \textcolor{green}{do his old job}.
\\
\hline
\end{tabular}
\end{table*}

\newpage
\begin{table*}[!ht]
\centering
\caption{Examples of sentiment steering combined with sentiment prompting applied to the Llama-3.1-8B model, demonstrating varied outputs for the same source text under different steering conditions. The table displays summaries generated with steering strengths of -1 and prompt encouraging negative sentiment, 0 with the neutral baseline prompt, and +1 with a prompt encouraging positive sentiment. Within these generated texts, \textcolor{red}{negative words and phrases are highlighted in red}, while \textcolor{green}{positive phrasings are highlighted in green} to visually illustrate the impact of steering on language characteristics.}
\label{tab:sentiment_examples_zlatan_burger}
\begin{tabular}{|p{0.33\textwidth}|p{0.25\textwidth}|p{0.33\textwidth}|}
\hline
\textbf{$\lambda = -1$ + negative sentiment prompting} & \textbf{$\lambda = 0$ + neutral prompting} & \textbf{$\lambda = 1$ + positive sentiment prompting} \\
\hline
The creation of \"Le Zlatan\" burger, a massive 600 gram burger, is a \textcolor{red}{cynical attempt to capitalize} on Zlatan Ibrahimovic's fame and reputation, with the French restaurateur Jean-Philippe Grandin \textcolor{red}{justifying the excessive price} by claiming Ibrahimovic's ego is \textcolor{green}{"extraordinary"}. This burger is a symbol of the \textcolor{red}{inflated ego and inflated prices} that come with celebrity worship, and is likely to be a turn-off for many diners. The burger's \textcolor{red}{outrageous price} of $41$ and its \textcolor{red}{stomach-busting contents} are a clear indication that Grandin is more interested in profiting from Ibrahimovic's fame than in serving quality food to his customers.
&
Zlatan Ibrahimovic's former team, Paris Saint-Germain, has a new burger in its honor, \"Le Zlatan\", which weighs over 1 kilogram and costs \$41. The massive burger, created by a French restaurateur, is a mixture of minced beef, onions, bacon, and three types of cheese. The burger is a tribute to Ibrahimovic, who has played for PSG and is known for his \textcolor{green}{impressive skills} on the field.
&
Zlatan Ibrahimovic's \textcolor{green}{culinary legacy lives on} through a \textcolor{green}{mouth-watering, 600-gram burger} that has been crafted by a French restaurateur in honor of the Swedish striker. The \"Le Zlatan\" burger is a \textcolor{green}{testament to Ibrahimovic's status as a formidable player}, weighing in at over one kilogram and featuring a blend of minced beef, onions, bacon, and three types of cheese. As a fan of the team, the restaurateur has created a \textcolor{green}{unique and extraordinary burger} that \textcolor{green}{embodies the excellence and supremacy} of Ibrahimovic, and with the player extending his contract until 2016, it's clear that this burger will \textcolor{green}{remain a culinary icon for years to come}.
\\
\hline
\end{tabular}
\end{table*}

\newpage
\subsection{Impact across Model Sizes}
\label{app:impact_of_model_scale}
The impact of prompting across model scales can be found in Appendix \ref{app:prompting_effects_across_model_scales}.

\noindent The impact of combined steering and prompting across model scales can be found in Appendix \ref{app:combined_steering_and_prompting_effects_across_model_scales}.

\noindent Some examples for summaries generated by model of different sizes can be found in Appendix \ref{app:individual_samples}.

\subsection{Impact of Steering on Summary Faithfulness and Hallucinations}
\label{app:hallucinations}

To further investigate the efficacy-quality trade-off observed in our automated metrics (ROUGE, BERTScore), we analyzed the impact of steering vectors on factual faithfulness. We posit that steering introduces a risk of ``attribute-congruent hallucinations,'' where the model fabricates information to satisfy the steering direction, particularly when the target attribute conflicts with the ground truth of the source text (e.g., steering for positive sentiment on an overwhelmingly negative news article).

To quantify this risk, we conducted a human evaluation on summaries from the NEWTS dataset. We randomly selected 25 samples and evaluated hallucination rates at three steering strengths ($\lambda \in \{-1.5, 0, 1.5\}$) for two distinct properties: \textbf{Sentiment} (a content-heavy attribute) and \textbf{Readability} (a stylistic attribute). We compared performance across two model sizes: Llama-3.2-1B and Llama-3.1-8B. The results, presented in Table~\ref{tab:hallucination_study}, confirm that steering generally increases hallucination rates compared to the unsteered baseline ($\lambda=0$).

\begin{table}[htp]
    \centering
    \caption{Number of hallucinations (contradictions to source) observed in 25 randomly selected samples per setting. Steering generally increases the rate of hallucinations, particularly for content-heavy attributes like sentiment.}
    \label{tab:hallucination_study}
    \begin{tabular}{l c c c c}
        \toprule
        \textbf{Model} & \textbf{Property} & $\mathbf{\lambda = -1.5}$ & $\mathbf{\lambda = 0}$ (Baseline) & $\mathbf{\lambda = +1.5}$ \\
        \midrule
        Llama-3.2-1B & Sentiment & 10 & 4 & 9 \\
        Llama-3.2-1B & Readability & 7 & 4 & 5 \\
        \midrule
        Llama-3.1-8B & Sentiment & 3 & 1 & 4 \\
        Llama-3.1-8B & Readability & 2 & 1 & 1 \\
        \bottomrule
    \end{tabular}
\end{table}

We observe three key trends:
\begin{itemize}
    \item \textbf{Content vs. Style:} Steering for sentiment, which requires altering the semantic content of the summary, resulted in significantly higher hallucination rates than steering for readability. This supports the hypothesis that when the steering target (e.g., ``positive sentiment'') contradicts the source text (e.g., ``negative news''), the model is forced to hallucinate details to resolve the conflict.
    \item \textbf{Steering Increases Risk:} For the smaller model (1B), steering in either direction (positive or negative) increased the number of contradictions compared to the baseline.
    \item \textbf{Model Size Robustness:} The larger Llama-3.1-8B model exhibited fewer hallucinations overall (lower baseline) and was more robust to steering-induced fabrications compared to the 1B model.
\end{itemize}

\end{document}